\documentclass[12pt]{article}

\usepackage{mathptmx}

\usepackage{graphicx}

\usepackage[letterpaper,margin=1in]{geometry}

\renewenvironment{abstract}
	{\quotation}
	{\endquotation}

\date{}

\makeatletter
\renewcommand{\fnum@figure}{\textbf{Figure \thefigure}}
\renewcommand{\fnum@table}{\textbf{Table \thetable}}
\makeatother

\usepackage{scicite}

\usepackage{url}

\usepackage{booktabs}
\usepackage{bm}
\usepackage{capt-of}
\usepackage{siunitx}
\usepackage{xspace}
\usepackage{xcolor}
\usepackage{multirow}
\usepackage{wrapfig}
\usepackage{algorithm}
\usepackage{algorithmicx,algpseudocode}
\usepackage{lmodern,babel,adjustbox,booktabs,multirow}
\usepackage{mathrsfs}
\usepackage{arydshln}
\usepackage{amsmath}
\usepackage{txfonts}

\makeatletter
\DeclareRobustCommand\onedot{\futurelet\@let@token\@onedot}
\def\@onedot{\ifx\@let@token.\else.\null\fi\xspace}
\def\eg{\emph{e.g}\onedot}
\def\ie{\emph{i.e}\onedot}

\def\etal{\emph{et al}\onedot}
\makeatother

\DeclareMathOperator*{\argmax}{arg\,max}
\DeclareMathOperator*{\argmin}{arg\,min}

\usepackage{color}
\newcommand{\revise}[1]{\textcolor{black}{#1}}

\def\scititle{
	FairDiffusion: Enhancing Equity in Latent Diffusion Models via Fair Bayesian Perturbation
}
\title{\bfseries \boldmath \scititle}

\author{
	Yan Luo$^{1,2,6\dagger}$,
	Muhammad Osama Khan$^{4\dagger}$,
        Congcong Wen$^{3,4\dagger}$, \\
	Muhammad Muneeb Afzal$^{4}$, 
     Titus Fidelis Wuermeling$^{1,2}$,
     Min Shi$^{1,2}$,
     Yu Tian$^{1,2}$,\\
     Yi Fang$^{3,4\mathsection}$,
     Mengyu Wang$^{1,2,5,6\ast\mathsection}$,
    \and
	\small$^{1}$Harvard AI and Robotics Lab, Schepens Eye Research Institute of Massachusetts Eye and Ear,\\
 \small Harvard Medical School, Boston, MA 02114 USA.\and
	\small$^{2}$Harvard Ophthalmology AI Lab, Schepens Eye Research Institute of Massachusetts Eye and Ear,\\
 \small Harvard Medical School, Boston, MA 02114, USA.\and
	\small$^{3}$Embodied AI and Robotics (AIR) Lab, New York University, 6 MetroTech Center, Brooklyn, 11201, NY, USA.\and
 \small$^{4}$Center for Artificial Intelligence and Robotics, New York University Abu Dhabi,\\
 \small Saadiyat Island, Abu Dhabi, 129188, United Arab Emirates.\and
 \small$^{5}$Kempner Institute for the Study of Natural and Artificial Intelligence, Harvard University, Boston, MA 02134, USA.\and
  \small$^{6}$Broad Institute of MIT and Harvard, Cambridge, MA 02142, USA.\and
	\small$^\ast$Corresponding author: Mengyu Wang. Email: mengyu\_wang@meei.harvard.edu\and
	\small$^\dagger$These authors contributed equally as co-first authors.\and
    \small$^\mathsection$These authors contributed equally.
}

\begin{document} 

\maketitle

\begin{abstract} \bfseries \boldmath
\revise{Recent advancements in generative AI, particularly diffusion models, have proven valuable for text-to-image synthesis. In healthcare, these models offer immense potential in generating synthetic datasets and aiding medical training. However,  despite these strong performances, it remains uncertain if the image generation quality is consistent across different demographic subgroups. To address this, we conduct a comprehensive analysis of fairness in medical text-to-image diffusion models. Evaluations of the Stable Diffusion model reveal substantial disparities across gender, race, and ethnicity. To reduce these biases, we propose FairDiffusion, an equity-aware latent diffusion model that improves both image quality and the semantic alignment of clinical features. Additionally, we design and curate FairGenMed, a dataset tailored for fairness studies in medical generative models. FairDiffusion is further assessed on HAM10000 (dermatoscopic images) and CheXpert (chest X-rays), demonstrating its effectiveness in diverse medical imaging modalities. Together, FairDiffusion and FairGenMed advance research in fair generative learning, promoting equitable benefits of generative AI in healthcare.}
\end{abstract}

\noindent





\section*{Introduction}
Generative modeling has witnessed rapid progress, powered largely by advancements in diffusion models in recent years.
Diffusion models \cite{ho2020denoising,sohl2015deep} are used as powerful tools in various scientific fields, offering innovative approaches to complex problems, such like protein structure discovery \cite{wu2024protein}, quantum circuit synthesis \cite{furrutter2024quantum}, peptide conformational sampling \cite{abdin2024direct}, molecular linker design \cite{igashov2024equivariant}, or embodied artificial agents \cite{berrueta2024maximum}.
Specifically, text-to-image diffusion models like Stable Diffusion~\cite{rombach2022high} have demonstrated \revise{remarkable} utility across a wide range of different domains. This text-conditioned generative modeling is especially useful in the medical domain, where it offers immense potential in generating synthetic datasets and training medical students to understand different conditions (e.g., eye diseases). By leveraging advanced image synthesis methods~\revise{\cite{pinaya2022brain,packhauser2023generation,madani2018chest,shah2022dc,motamed2021data,malygina2019data,karbhari2021generation,srivastav2021improved,moris2022unsupervised, song2021solving, xie2022measurement, chung2022score, hu2022unsupervised, ozbey2023unsupervised}},
medical educators can create highly realistic and diverse datasets of various eye disease conditions. These synthesized images can accurately represent a wide range of pathological manifestations, including subtle variations and rare cases that may be challenging to encounter in real-world clinical settings, resulting in more skilled medical practitioners.

However, despite these strong performances, it remains unclear whether the quality of image generation is consistent across different demographic subgroups. Fairness gaps in image generation quality across different protected attributes raise \revise{critical} ethical and equity issues, especially in the medical domain where such biases could inadvertently perpetuate healthcare disparities~\revise{\cite{xu2024addressing, tian2025fairdomain, ktena2024generative, drukker2023toward}}. Hence, ensuring fairness in medical image generation is not only a critical challenge from a technical standpoint but also a moral imperative in order to ensure that advances in generative models do not exacerbate existing healthcare inequalities.

To address these critical concerns, we conduct the first comprehensive study on the fairness of medical text-to-image diffusion models. Specifically, using the popular Stable Diffusion~\cite{rombach2022high} model, we conduct two types of fairness evaluations. Firstly, we evaluate the generation quality of the diffusion model across different demographic subgroups. Secondly, we train a classifier with the generated dataset to examine the semantic correlation of clinical features in text prompts with the generated images across various subgroups. Our comprehensive evaluations of the Stable Diffusion model reveal \revise{notable} biases across all protected attributes in terms of both image generation quality as well as semantic correlation of clinical features. For instance, the Female, \revise{white}, and non-Hispanic subgroups show improved image generation performance across the protected attributes of gender, race, and ethnicity, whereas the Male, Asian, and non-Hispanic subgroups are favored in the semantic correlation setting on the glaucoma classification task.

\begin{figure*}[t]
  \centering
    \includegraphics[width=0.5\textwidth]{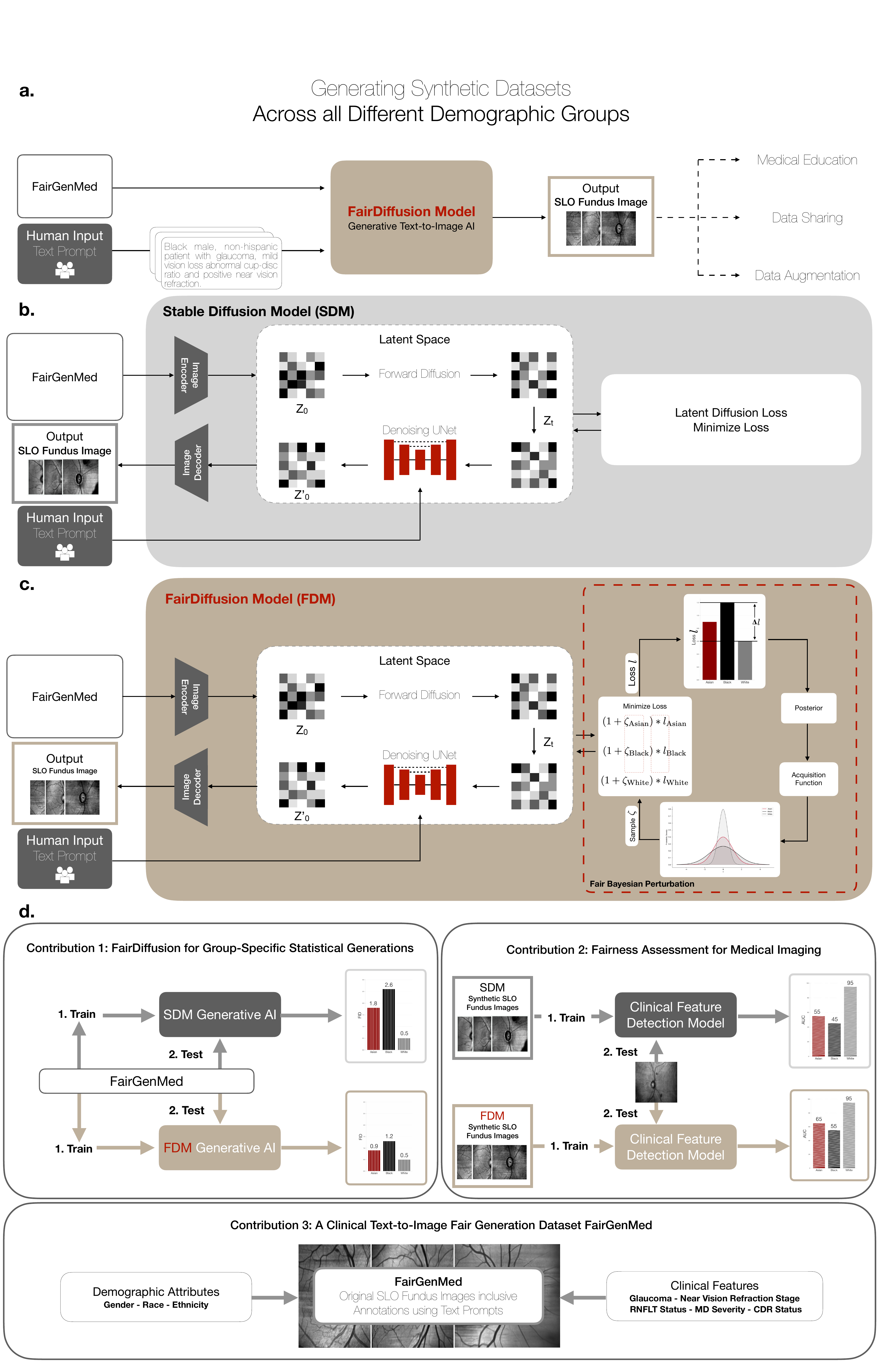}
  \caption{\revise{\textbf{Overview of FairDiffusion Framework for Equitable Medical Image Generation.}}\textbf{a.} General workflow of SLO fundus images generation with generative artificial intelligence (AI) models from text prompt instructions describing clinical features. A carefully designed and curated clinical dataset FairGenMed is used to perform the text-to-image generation task in this study. The generative AI can be potentially used for the purposes of medical education, data sharing, and data augmentation. \textbf{b.} The off-the-shelf generative AI model termed Stable Diffusion model uses clinical features as text prompt input to generate corresponding SLO fundus images.
  \textbf{c.} The FairDiffusion model uses clinical features as text prompt input to generate SLO fundus images with fair Bayesian perturbation to determine demographic group-specific statistical generation parameters. 
  \textbf{d.} Contribution 1: FairDiffusion for group-specific statistical generations improves generative AI equity. Contribution 2: Fairness assessment pipeline for medical imaging to quantify the semantic correlations of clinical features between text prompt input and generated SLO fundus images. Contribution 3: A clinical text-to-image fair generation dataset FairGenMed provides a unique dataset for studying fairness issues in generative AI for medical tasks.
  }
  
   \label{fig:tease_img}
 \end{figure*}

To tackle these fairness gaps, we propose FairDiffusion (Fig. \ref{fig:tease_img}), an equity-aware stable diffusion model designed to improve fairness in medical image generation. FairDiffusion employs a Bayesian Optimization based approach and works by adaptively perturbing the learning process with samples from each demographic group in order to achieve fair generative learning. \revise{Through extensive experiments across three medical imaging modalities, including SLO fundus images, dermatoscopic images, and chest X-rays, we demonstrate that our proposed FairDiffusion method substantially improves both overall performance and fairness across image generation quality as well as semantic correlation of clinical features.} \revise{Using results from the ophthalmology SLO fundus dataset as an example,} in the image generation quality setting, FairDiffusion leads to notable improvements for the Black and Hispanic subgroups, with \revise{Fréchet Inception Distance (FID)} gains of 7.84 and 11.79 respectively. In the semantic correlation setting, FairDiffusion achieves large gains for the Asian and Male subgroups, with AUC enhancements of 10.03 and 7.58 respectively, on the glaucoma classification task. \revise{Consistent improvements are also observed across dermatoscopic images and chest X-rays, further underscoring FairDiffusion's generalizability and effectiveness in addressing fairness concerns across diverse medical imaging modalities.}

In addition to the proposed FairDiffusion method, we also introduce FairGenMed, the first dataset for studying fairness of medical generative models. While existing radiology datasets could be repurposed for fairness analysis, prior works~\cite{seyyed2020chexclusion,irvin2019chexpert} have shown the difficulties of using these datasets for fairness studies. These challenges stem from the automatic extraction of~\textit{ground-truth} labels from radiology reports, potentially resulting in inaccurate fairness assessments due to the noisy labels. In contrast to existing datasets, FairGenMed not only provides ground-truth labels but also includes detailed quantitative measurements of different clinical conditions such as cup-disc ratio (CDR), retinal nerve fiber layer thickness (RNFLT), and near vision refraction (NVR). Since these conditions are naturally correlated with retinal anatomy and pathophysiology, they serve as excellent resources for studying the semantic correlation between text prompts and anatomical regions across various demographic subgroups. With the rapid increase in popularity of generative models, FairDiffusion and FairGenMed have \revise{promising} potential to drive advancements in equitable and fair generative learning algorithms.

\begin{figure*}[t]
  \centering
    \includegraphics[width=0.9\textwidth]{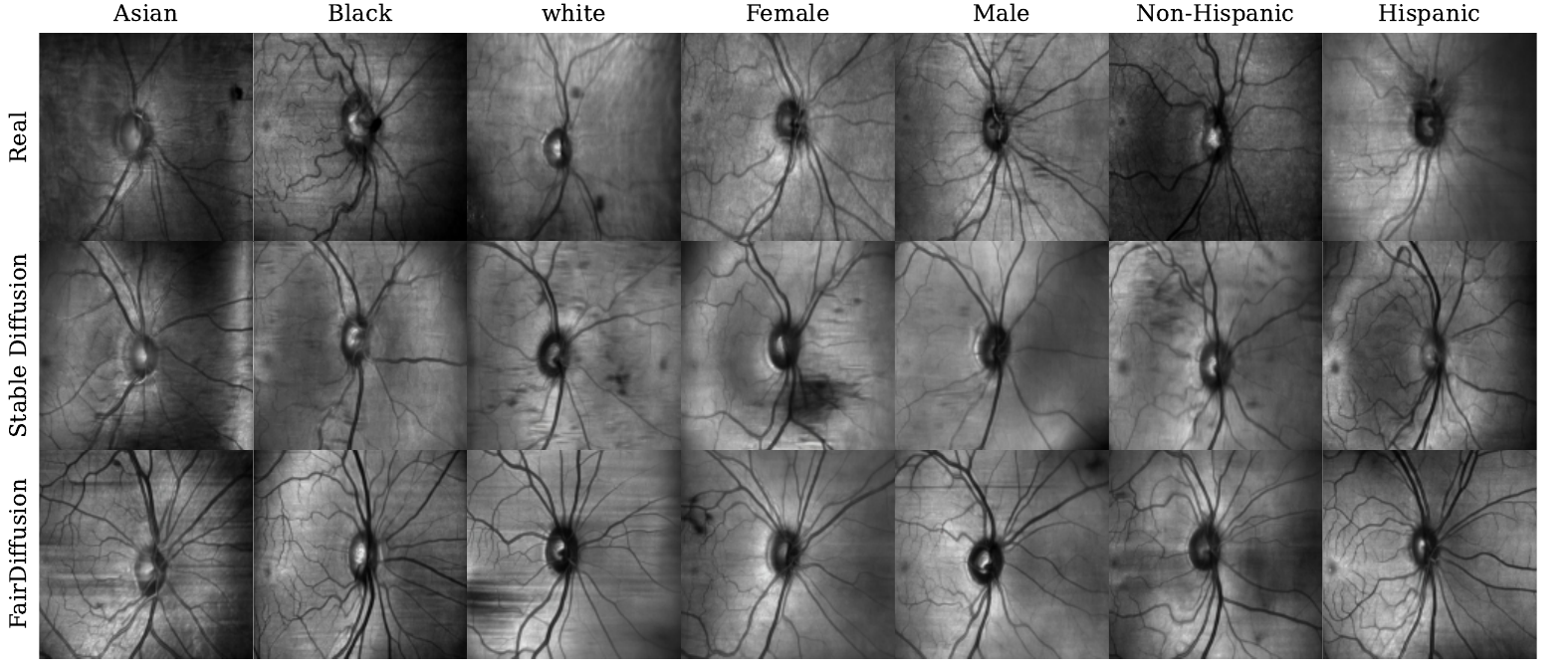}
  \caption{\revise{\textbf{Qualitative visualizations of Stable Diffusion and FairDiffusion generated images against real images from the FairGenMed dataset.} Given the images in this figure, FairDiffusion achieves a IS score of 1.5, while Stable Diffusion achieves a IS score of 1.3.}
  }
  \label{fig:qual_vis}
\end{figure*}

\section*{Results}
We train a FairDiffusion model using the proposed Fair Bayesian Perturbation approach. For comparison, we benchmark against the widely-used Stable Diffusion model, following the same training setup for both methods to ensure a fair evaluation. Both models are trained with 7,000 Scanning Laser Ophthalmoscopy (SLO) fundus images from the proposed FairGenMed dataset, where each patient is associated with only one SLO fundus image. The study workflow is delineated in Fig.~\ref{fig:tease_img}. Our approach enhances fairness across three key dimensions. Firstly, we improve fairness in image generation quality across various protected attributes such as gender, race, and ethnicity (Fig.~\ref{fig:tease_img}d: Contribution 1). Secondly, we ensure that the semantic correlation of clinical features in text prompts with the generated retinal scans is consistent across various demographic subgroups. This is accomplished via training classifiers with the generated images and comparing the classification performance on different clinical features across these subgroups (Fig.~\ref{fig:tease_img}d: Contribution 2). Moreover, we also release FairGenMed, the first dataset for studying fairness of medical generative models that provides patients' SLO fundus images, diagnosis and detailed measurements of multiple clinical features (Fig.~\ref{fig:tease_img}d: Contribution 3), which are used to construct the text prompts. In addition to these quantitative performances measured via the equity-scaled metrics (ES-FID, ES-IS, ES-AUC), we also conduct comprehensive qualitative analyses, demonstrating that FairDiffusion consistently outperforms Stable Diffusion in training fair generative models.
\revise{To assess the generalizability of the proposed FairDiffusion across diverse medical imaging modalities, we further conduct evaluations on two external datasets: HAM10000 (dermatoscopic images) and CheXpert (chest X-rays), focusing on both image generation and classification tasks.}



\begin{figure}[t]
  \centering
    \includegraphics[width=0.8\textwidth]{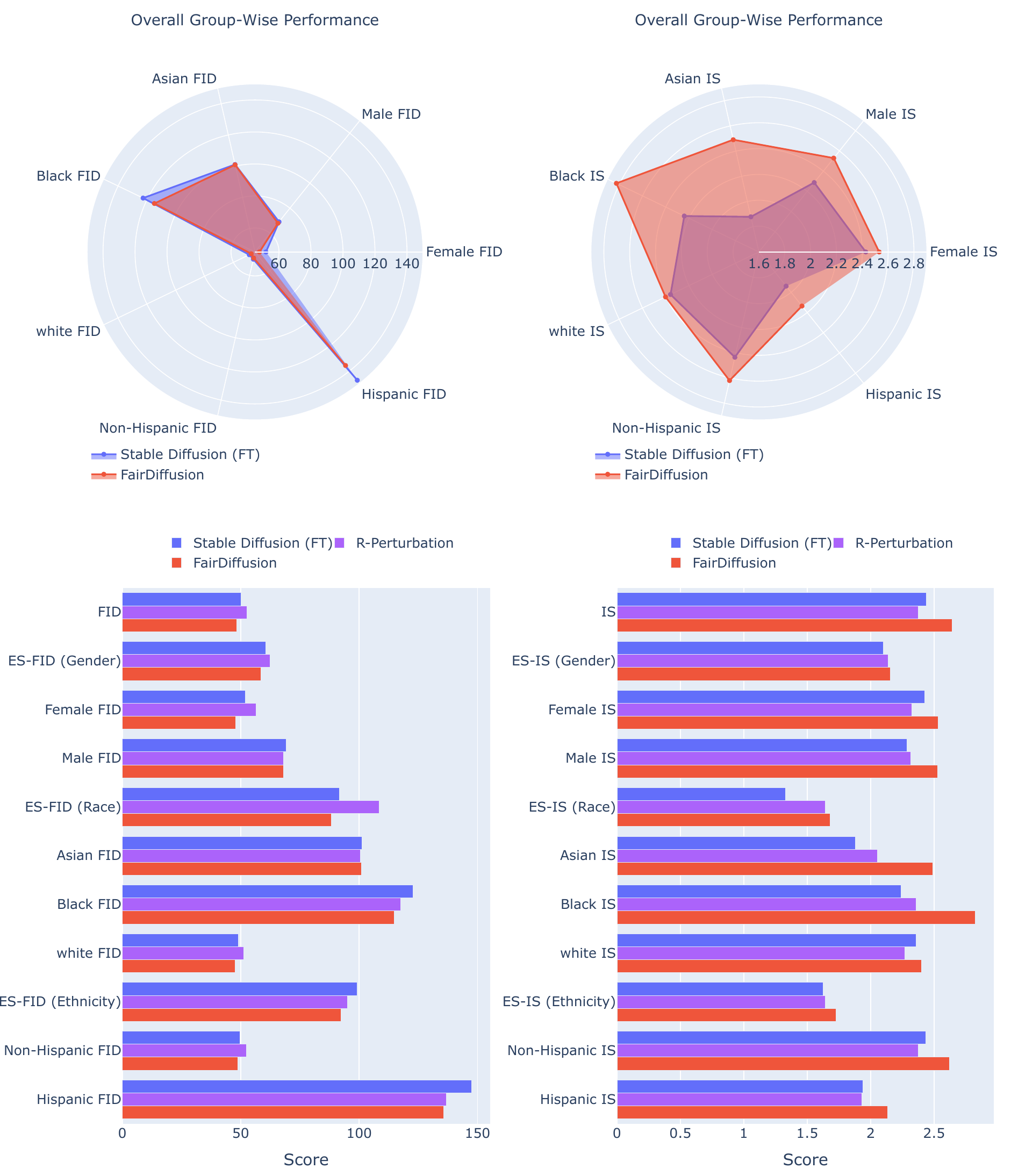}
  \caption{\revise{\textbf{Fairness Evaluation of Image Generation Quality.}} We present a comprehensive comparison of performance metrics for SLO fundus image generation across different models and demographic attributes. The left column shows FID scores, while the right column displays IS results. Lower FID and higher IS indicate better performance. Performance is broken down by overall group-wise metrics and ES-FID and ES-IS metrics for gender, race, and ethnicity. The ES measures the consistency of performance across different demographic subgroups, with lower scores indicating more equitable performance.
  This visualization allows for a detailed analysis of each model's effectiveness and fairness in generating SLO fundus images across diverse patient demographics.
  }
  \label{fig:gen_plot}
\end{figure}

\begin{figure}[t]
  \centering
    \includegraphics[width=0.8\textwidth]{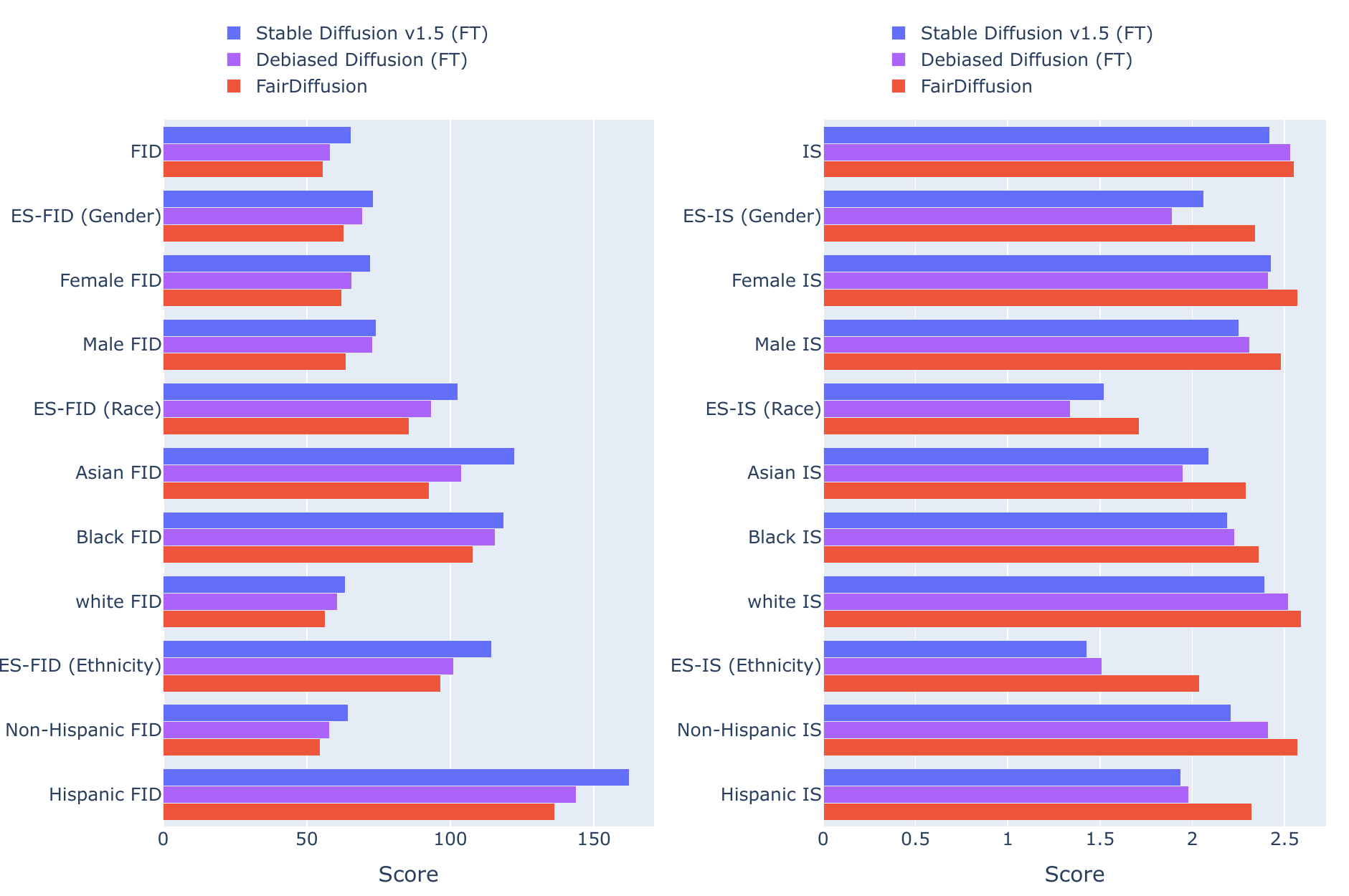}
  \caption{\revise{\textbf{Generation performance of Stable Diffusion v1.5, debiased diffusion, and FairDiffusion.}} Left: FID scores for different demographic subgroups and their equity-scaled metrics (ES-FID). Lower FID scores indicate better image quality. Right: IS scores and their equity-scaled metrics (ES-IS) across demographic groups. Higher IS scores indicate better generation quality. In this experiments, following \cite{shen2024finetuning}, debiased diffusion and FairDiffusion are both based on Stable Diffusion v1.5.
  }
  \label{fig:sd15_gen_plot}
\end{figure}


\begin{figure}[t]
  \centering
    \includegraphics[width=0.8\textwidth]{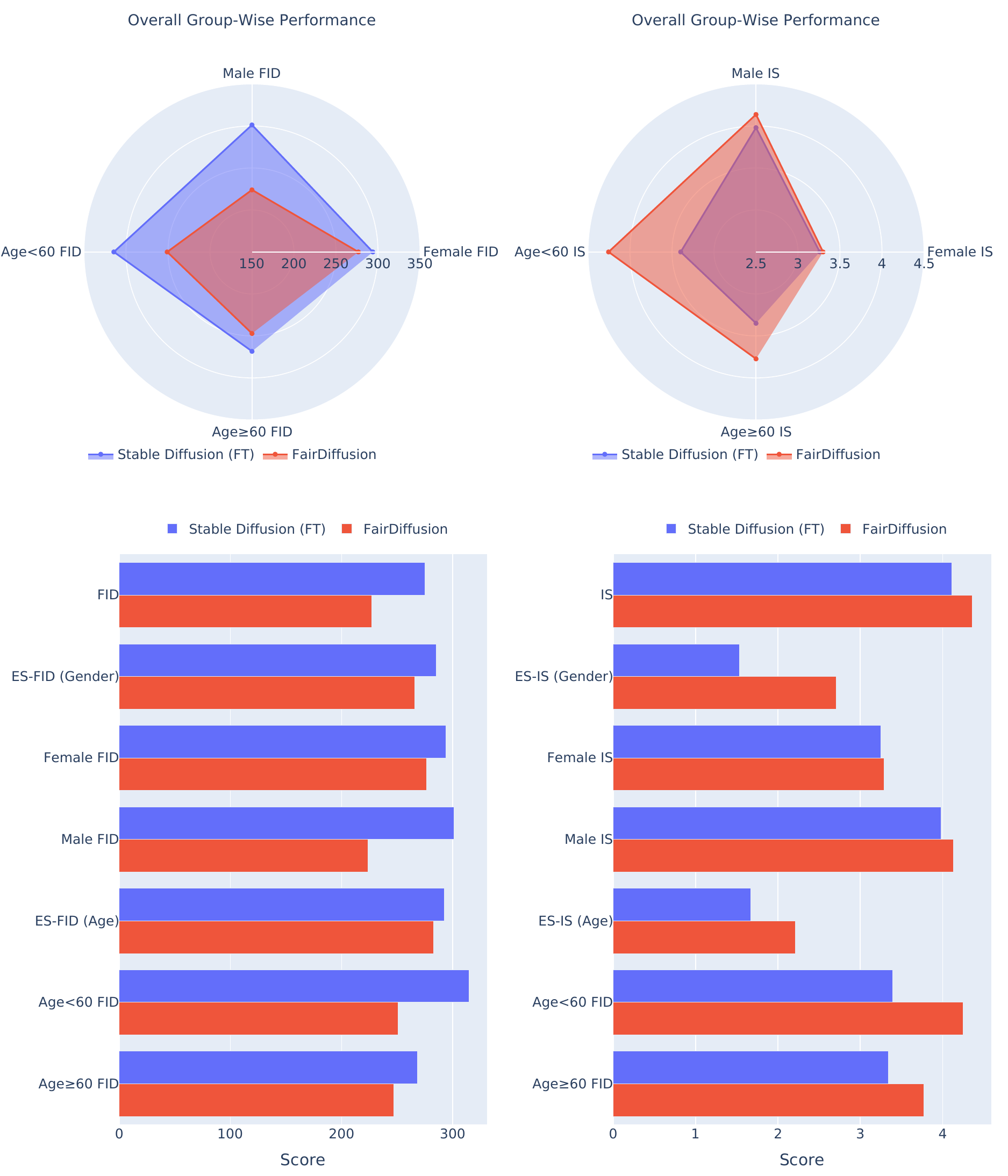}
  \caption{\revise{\textbf{Fairness Evaluation of Image Generation Quality on the HAM10000 dataset.}} We present a comprehensive comparison of performance metrics for dermatoscopic image generation across different models and demographic attributes. The left column shows FID scores, while the right column displays IS results. Lower FID and higher IS indicate better performance. Performance is broken down by overall group-wise metrics and ES-FID and ES-IS metrics for age and gender. The ES measures the consistency of performance across different demographic subgroups, with lower scores indicating more equitable performance. This visualization allows for a detailed analysis of each model's effectiveness and fairness in generating dermatoscopic images across diverse patient demographics.
  }
  \label{fig:gen_plot_ham}
\end{figure}

\begin{figure}[t]
  \centering
    \includegraphics[width=0.8\textwidth]{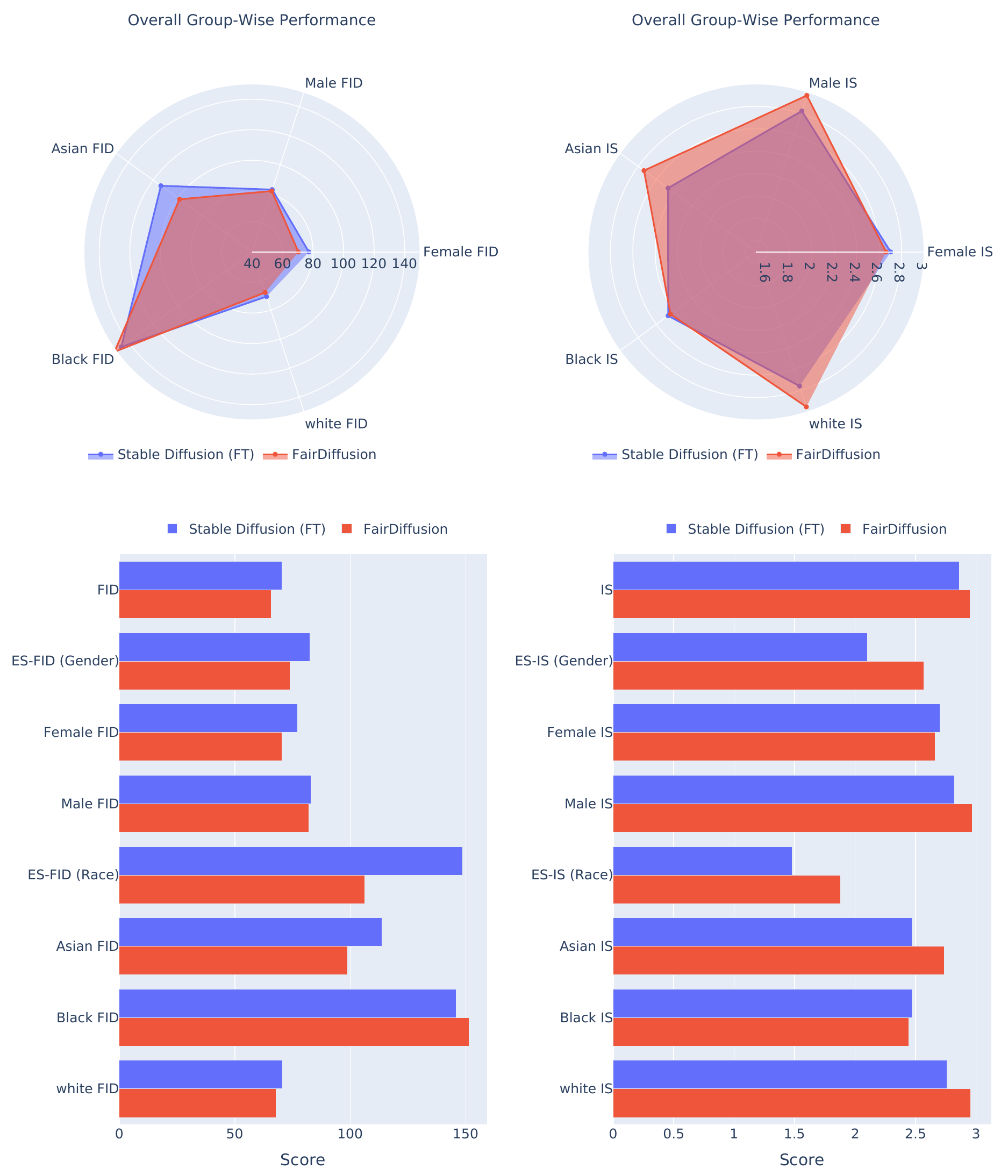}
  \caption{\revise{\textbf{Fairness Evaluation of Image Generation Quality on the CheXpert dataset.}} We present a comprehensive comparison of performance metrics for chest radiograph image generation across different models and demographic attributes. The left column shows FID scores, while the right column displays IS results. Lower FID and higher IS indicate better performance. Performance is broken down by overall group-wise metrics and ES-FID and ES-IS metrics for gender and race. The ES measures the consistency of performance across different demographic subgroups, with lower scores indicating more equitable performance. This visualization allows for a detailed analysis of each model's effectiveness and fairness in generating chest radiograph images across diverse patient demographics.
  }
  \label{fig:gen_plot_chexpert}
\end{figure}

\begin{figure*}[t]
  \centering
    \includegraphics[width=1.\textwidth]{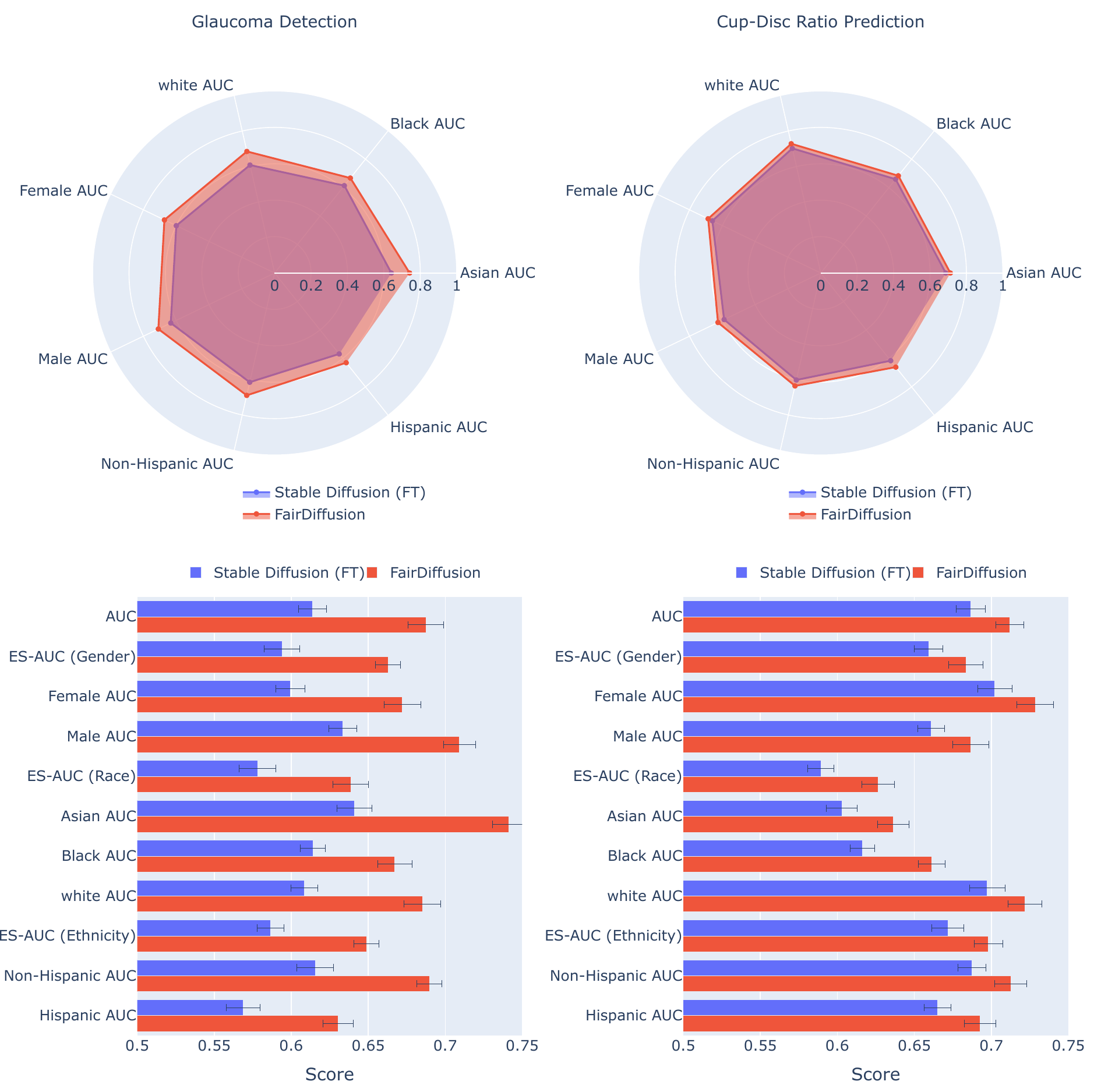}
  \caption{\revise{\textbf{Fairness Evaluation of Semantic Correlation in Clinical Feature Classification.}} We present a comprehensive fairness evaluation of the semantic correlation of clinical features across different demographic groups, focusing on two key classification tasks: Glaucoma Detection (left) and Cup-Disc Ratio Prediction (right).
  It compares the performance of deep learning model trained with the images generated by Stable Diffusion (Fine-Tuned) and FairDiffusion. The results are organized by race (Asian, Black, \revise{white}), gender (Female, Male), and ethnicity (Non-Hispanic, Hispanic), showcasing both overall group-wise performance and equity scores. This visualization enables a nuanced analysis of each model's effectiveness and fairness in classifying critical clinical features across patient demographics, highlighting the delicate balance between performance and equity in medical image analysis tasks.
  }
  \label{fig:cls_plot}
\end{figure*}


\begin{figure}[t]
  \centering
    \includegraphics[width=1.0\textwidth]{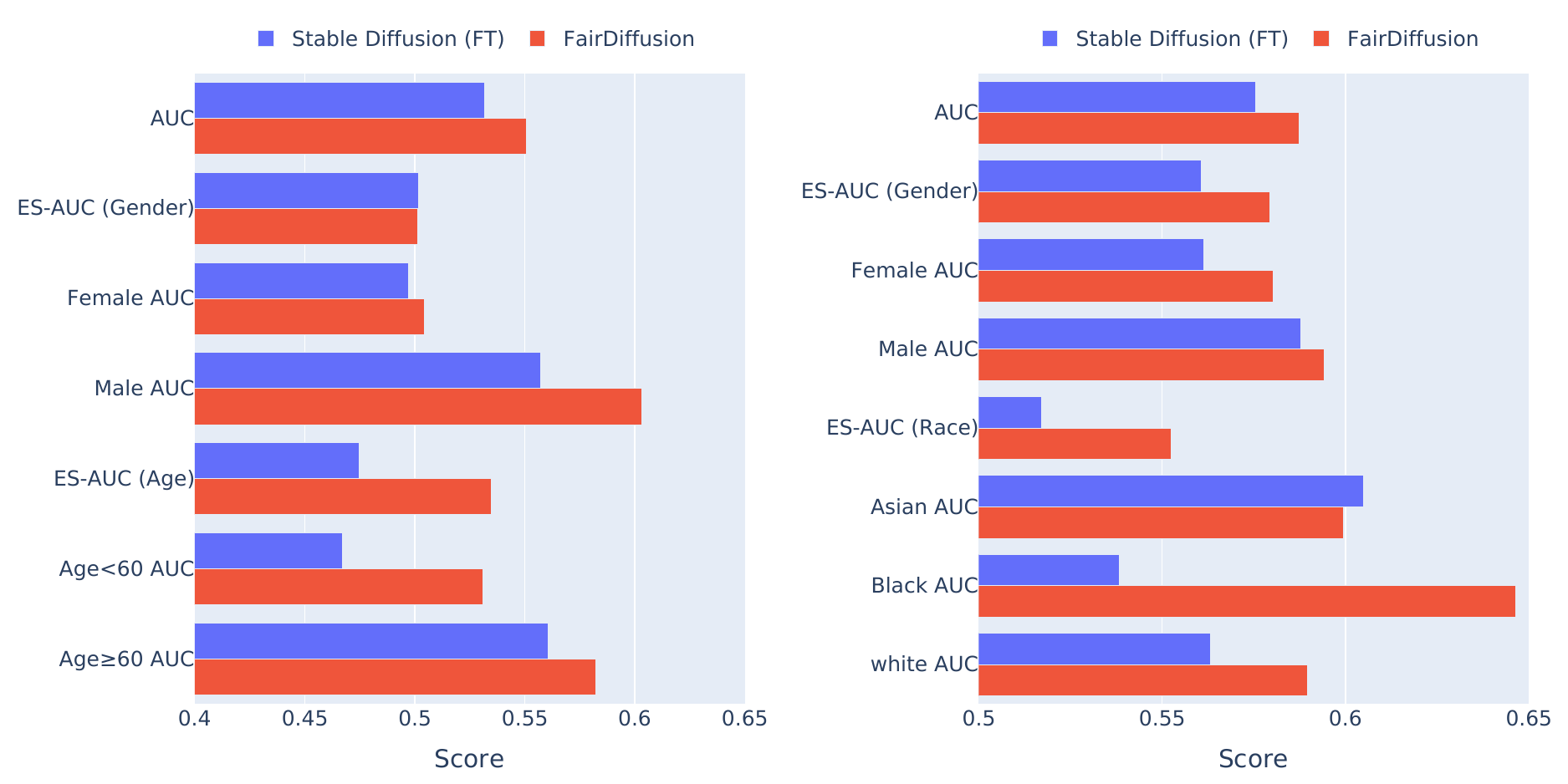}
  \caption{\revise{\textbf{Fairness Evaluation of Semantic Correlation in Clinical Feature Classification on HAM10000 and CheXpert Datasets.}} We present a comprehensive fairness evaluation of the semantic correlation of clinical features across different demographic groups, focusing on two key classification tasks: Skin Cancer Detection (left) and Pleural Effusion Detection (right). It compares the performance of deep learning model trained with the images generated by Stable Diffusion (Fine-Tuned) and FairDiffusion.  This visualization enables a nuanced analysis of each model’s effectiveness and fairness in classifying critical clinical features across patient demographics, highlighting the delicate balance between performance and equity in medical image analysis tasks.
  }
  \label{fig:gen_plot_cls_ham_chexpert}
\end{figure}

\begin{figure*}[t]
  \centering
    \includegraphics[width=.8\textwidth]{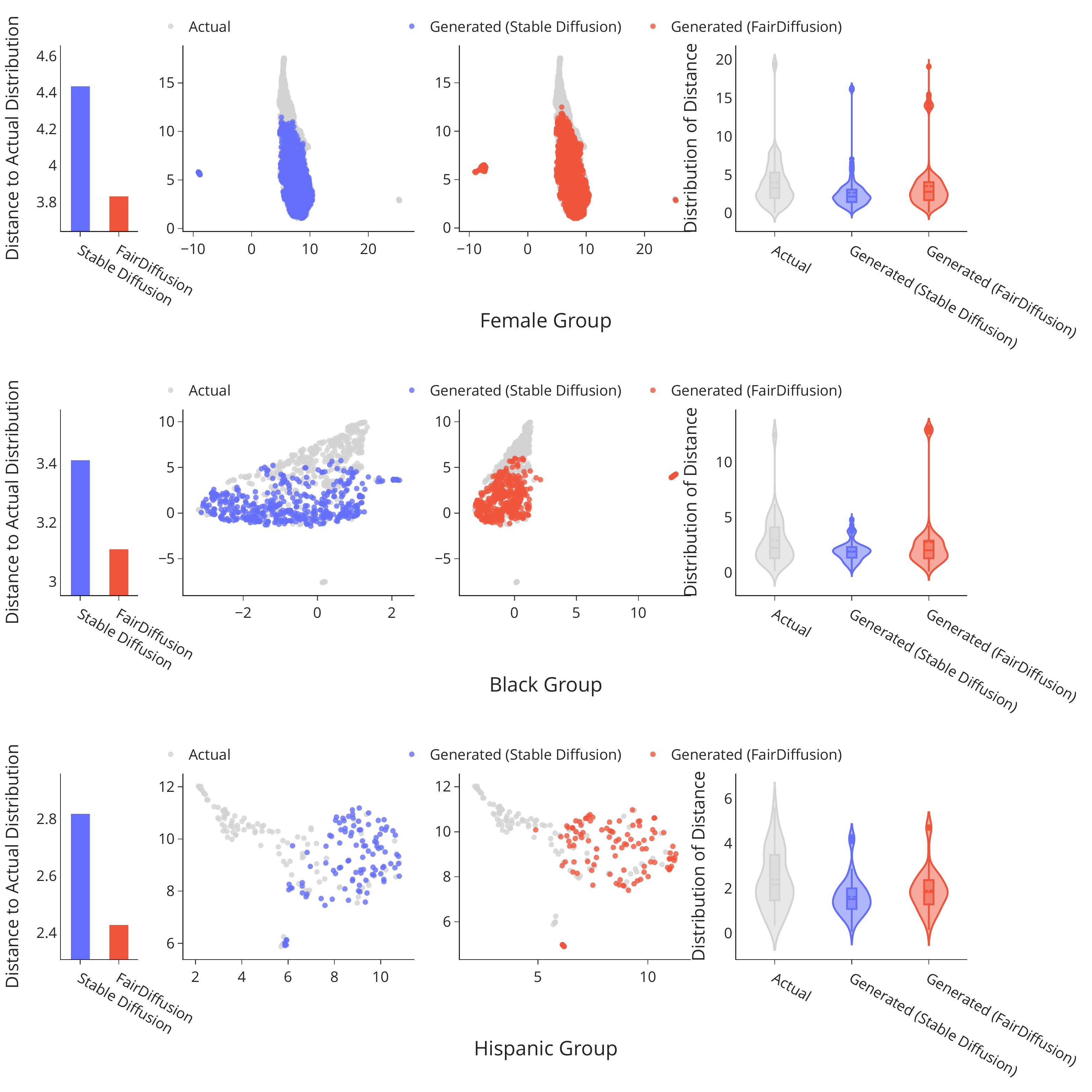}
  \caption{\revise{\textbf{UMAP Analysis of Image Generation Models for Underrepresented Groups, i.e., Female, Black, and Hispanic.}} We present the UMAP (Uniform Manifold Approximation and Projection) analysis comparing the performance of Stable Diffusion and FairDiffusion models in generating images for three underrepresented groups. The analysis is divided into four components for each group: (first column) Sliced Wasserstein Distance between generated and actual images, (second column) 2D UMAP distribution of Stable Diffusion-generated images compared to actual images, (third column) 2D UMAP distribution of FairDiffusion-generated images compared to actual images, and (fourth column) distribution of distances to each group's barycenter. This multi-faceted visualization enables a detailed comparison of how accurately each model captures the characteristics of underrepresented groups in generated images, revealing differences in distribution and proximity to actual image data across demographic categories. By juxtaposing these visualizations, it provides insights into the relative performance and fairness of the two generation models in representing diverse populations.
  }
  \label{fig:umap}
\end{figure*}

\subsection*{Qualitative Visualizations of Stable Diffusion vs FairDiffusion}
We compare generated images of the trained Stable Diffusion and FairDiffusion models against real SLO fundus images from the FairGenMed dataset. Fig.~\ref{fig:qual_vis} illustrates the visualizations, demonstrating that both models generate realistic-looking retinal scans. However, since it is difficult to visually compare the generated results, we design two evaluation pipelines for assessing the performance and fairness of these generative models. Firstly, we generate SLO fundus images across different demographic subgroups and compute the ES-FID and ES-IS metrics that give a comprehensive overview of fairness in terms of image generation quality. Secondly, in the healthcare context, it is also critical to assess the semantic correlation of clinical features mentioned in text prompts with the generated SLO fundus images. To this end, we design an evaluation pipeline that trains classifiers for various clinical features (e.g., glaucoma, cup-disc ratio) on generated SLO fundus images and measures the resulting diagnostic performance across various demographic subgroups on the real test set from the FairGenMed dataset. Next, we discuss these two fairness evaluations in detail.

\subsection*{Fairness in Image Generation Quality}
We conduct a comprehensive evaluation of the proposed FairDiffusion model against Stable Diffusion in terms of image generation quality. Particularly, Fig.~\ref{fig:gen_plot} compares Stable Diffusion (SD) with FairDiffusion both in terms of overall generation performance (50.1 vs. 48.3 on FID with $p<0.001$) as well as the subgroup fairness (e.g., 96.1 vs. 88.27 on Race ES-FID with $p<0.001$) across the three protected attributes of gender, race, and ethnicity. Moreover, we also report the group-wise FID and group-wise IS in order to understand the generation performance for the individual subgroups. 

Across the protected attribute of gender, we observe that the Stable Diffusion model generations are superior (51.8 on FID) for the Female subgroup compared to the Male subgroup (69.0 on FID). Across race, we observe that the generation performance is best for the \revise{white} subgroup followed by the Asian and Black subgroups. Lastly, across ethnicity, Stable Diffusion exhibits better generations for the non-Hispanic subgroup than the Hispanic subgroup. Whereas some of these discrepancies in generation performances could be attributed to the size imbalances in the training dataset (e.g., number of non-Hispanic patients $\ll$ number of Hispanic patients), this is unlikely to be the sole reason for these biases. For instance, the generative performance for the Asian subgroup is better than that for the Black subgroup despite the former containing the least number of training examples. Overall, we observe that the Stable Diffusion model exhibits \revise{notable} biases, favoring the Female, \revise{white}, and non-Hispanic subgroups across the protected attributes of gender, race, and ethnicity respectively.

To address these biases, we propose FairDiffusion -- a Bayesian Optimization based approach that encourages fair generation performance across the different subgroups. As illustrated in Fig.~\ref{fig:gen_plot}, FairDiffusion consistently outperforms ($p<0.001$) Stable Diffusion not only in terms of overall generation performance (48.3 vs. 50.1 on FID, 2.64 vs. 2.43 on IS) but also subgroup fairness (e.g., 88.2 vs. 91.6 on Race ES-FID, 1.67 vs. 1.33 on Race ES-IS, with $p<0.001$). Lower FID scores are better while higher IS scores are better.
\revise{In terms of the individual subgroup performance for the Female subgroup, FairDiffusion reduces the FID score from 51.8 to 48.0 ($p<0.001$) and improves the IS score from 2.43 to 2.64 ($p<0.005$). The Black subgroup sees a substantial FID reduction from 122.6 to 114.8 ($p<0.001$) and an IS improvement from 2.2 to 2.8 ($p<0.001$). In addition, the notable improvement is observed in the Hispanic subgroup, where FairDiffusion achieves an FID reduction from 1147.5 to 135.7 ($p<0.001$) and increases the IS score from 1.93 to 2.2 ($p<0.001$).}
Hence, our proposed FairDiffusion method not only improves the overall generation performance but also successfully alleviates the fairness gaps across all protected attributes. For reference, we also provide comparison against a random perturbation baseline (\textit{R-Perturbation}), which perturbs the learning process using random noise generated from a unit Gaussian distribution $\mathcal{N}(0,1)$. However, this fails to improve fairness, further validating the effectiveness of our Bayesian Optimization based learning method.
\revise{Also, as shown in Fig. \ref{fig:md_fid_is}, FairDiffusion achieves lower maximum discrepancy compared to both Stable Diffusion (FT) and R-Perturbation baselines, indicating more consistent generation quality across different demographic groups.}

\revise{
For a comprehensive evaluation, we present the results of FairDiffusion, built upon Stable Diffusion v1.5, compared against Stable Diffusion v1.5 and Debiased Diffusion \cite{shen2024finetuning}. As shown in Fig. \ref{fig:sd15_gen_plot}, FairDiffusion demonstrates superior performance across all metrics. It achieves better FID scores across all protected attributes (48.3 vs. 50.1 overall FID with $p<0.001$), with \revise{notable} improvements in fairness metrics (ES-FID of 88.2 vs. 96.1 for race, 91.3 vs. 94.8 for gender, and 92.5 vs. 97.2 for ethnicity with $p<0.001$). The improvements are most pronounced for historically underrepresented groups, with FID reductions of 7.84 for the Black subgroup (from 71.2 to 63.4) and 11.79 for the Hispanic subgroup (from 138.4 to 126.6). The IS metrics show similar trends, with FairDiffusion achieving higher overall IS (2.64 vs. 2.43) and improved equity-scaled metrics (ES-IS of 1.67 vs. 1.33 for race, 1.82 vs. 1.45 for gender, and 1.74 vs. 1.38 for ethnicity with $p<0.001$).}

\revise{To validate the generalizability of the proposed method across diverse medical imaging modalities, we further evaluate FairDiffusion on two additional medical imaging datasets, HAM10000 (dermatoscopic images) and CheXpert (chest X-rays). On HAM10000, as shown in Fig.~\ref{fig:gen_plot_ham}, FairDiffusion demonstrates substantial improvements in fairness metrics, with ES-FID reduced from 285.26 to 265.97 for the gender group and from 292.57 to 282.70 for the age group. Moreover, the model achieves higher ES-IS values, increasing from 1.53 to 2.71 for the gender group and from 1.67 to 2.21 for the age group. On CheXpert, as shown in Fig.~\ref{fig:gen_plot_chexpert}, FairDiffusion exhibits similar advancements, reducing ES-FID from 82.43 to 73.8 for the gender group and from 148.52 to 106.2 for the race group. Additionally, ES-IS values improve from 2.1 to 2.57 for the gender group and from 1.48 to 1.88 for the race group. These findings, coupled with improvements in overall image generation quality, underscore FairDiffusion's ability to enhance fairness and achieve consistent performance across multiple medical imaging domains and demographic subgroups.
}

\subsection*{Fairness in Semantic Correlation of Clinical Features}
In addition to the image generation quality, we also assess the semantic correlation of clinical features mentioned in text prompts with the generated SLO fundus images. To this end, we present a detailed fairness evaluation comparing the performance of classifiers trained on generated images from FairDiffusion and Stable Diffusion. Specifically, Fig.~\ref{fig:cls_plot} illustrates results on glaucoma and cup-disc ratio classification tasks from the proposed FairGenMed dataset. In addition to the overall classification performance measured via AUC, we also report the fairness metric ES-AUC as well as the individual group-wise AUCs.

Considering the ES-AUC metric from Fig.~\ref{fig:cls_plot}, we observe that Stable Diffusion yields the best fairness results across gender subgroups (0.594 on Gender ES-AUC), and the worst across race subgroups (0.578 on Race ES-AUC) on the glaucoma classification task. In terms of the individual subgroups, Male, Asian, and non-Hispanic are the preferred categories across gender, race, and ethnicity with AUCs of 0.633, 0.641, and 0.615, respectively. On the other hand, considering the cup-disc ratio classification task from Fig.~\ref{fig:cls_plot}, we observe that Stable Diffusion is most fair across ethnicity subgroups (0.617 on Ethnicity ES-AUC) and least fair across race subgroups (0.589 on Race ES-AUC). Moreover, the group-wise analysis reveals that the Female, \revise{white}, and non-Hispanic subgroups are preferred across the protected attributes of gender, race, and ethnicity with AUCs of 0.702, 0.697, and 0.687, respectively. Overall, we note that Stable Diffusion demonstrates \revise{notable} biases across all protected attributes.

Our proposed FairDiffusion method not only addresses these fairness gaps but also helps improve the overall classification performance across both glaucoma and cup-disc ratio classification tasks (Fig.~\ref{fig:cls_plot}). On glaucoma classification, FairDiffusion witnesses large gains on all subgroups, with the gains being especially large on the Asian and Male subgroups (0.100 and 0.758 on AUC with $p<0.001$, respectively). Moreover, on cup-disc ratio classification, FairDiffusion yields appreciable benefits on the Black and Hispanic subgroups, with improvements of 0.045 ($p<0.001$) and 0.027 ($p<0.001$) on AUC respectively.

\revise{In addition to FairGenMed, we evaluate FairDiffusion and Stable Diffusion on the HAM10000 and CheXpert datasets in Fig.~\ref{fig:gen_plot_cls_ham_chexpert} to further validate its generalizability across medical imaging modalities in the classification task. On the HAM10000 dataset, FairDiffusion demonstrates notable improvements in overall classification performance, with the AUC increasing from 53.19 to 55.07. For gender-specific performance, the model enhances the AUC for the Female subgroup from 49.70 to 50.45 and for the Male subgroup from 55.71 to 60.31. Furthermore, FairDiffusion achieves \revise{promising} advancements in age-related fairness, as reflected by an increase in ES-AUC from 47.46 to 53.48. Similarly, on the CheXpert dataset, FairDiffusion improves both fairness and classification performance across gender and race subgroups. The overall AUC increases from 57.55 to 58.74, with ES-AUC for gender rising from 56.07 to 57.93 and ES-AUC for race increasing from 51.72 to 55.24. These results underscore FairDiffusion's capability to enhance fairness and overall classification performance across diverse medical imaging domains, further validating its utility in addressing biases in generative AI.}

In summary, our comprehensive analyses reveal \revise{notable} biases in the widely used Stable Diffusion model. To address these biases, we propose FairDiffusion, which \revise{remarkably} improves overall performance as well as fairness across both image generation as well as semantic correlation of clinical features. We provide additional quantitative results in the supplementary material, further validating the effectiveness of FairDiffusion.

\subsection*{Qualitative UMAP Analysis}
Beyond the quantitative fairness results, we also conduct a thorough qualitative analysis comparing the UMAP representations of the Stable Diffusion and FairDiffusion generated images to the UMAP representations of the real images. Fig.~\ref{fig:umap} illustrates the UMAP analyses across various demographic subgroups, including Female, Black, and Hispanic. As depicted in the analysis, FairDiffusion consistently outperforms Stable Diffusion across all demographic subgroups, as indicated by the reduced distance (e.g., from 4.4 to 3.8 on the Female group with $p<0.001$) between the generated and actual image distributions.

\section*{Discussion}
In this study, we show that our proposed FairDiffusion model, trained via Fair Bayesian Perturbation, successfully alleviates fairness gaps across image generation quality as well as semantic correlation of clinical features. While previous research has highlighted the potential of advanced generative models~\cite{chambon2022adapting,chambon2022roentgen,khader2022medical,pinaya2022brain,packhauser2023generation,madani2018chest,shah2022dc,motamed2021data,malygina2019data,karbhari2021generation,srivastav2021improved,moris2022unsupervised} in healthcare, their performance across diverse demographic subgroups remains largely unexplored. This gap is critical since understanding the performance of these models across various protected attributes is essential before they can be deployed in sensitive domains such as healthcare. To bridge this research gap, we conduct the first comprehensive study on the fairness of medical text-to-image diffusion models, using the widely-adopted Stable Diffusion model as a case study. Our findings reveal \revise{notable} biases, both in terms of image generation quality as well as semantic correlation of clinical features. To mitigate these biases, we develop FairDiffusion, which enhances both overall performance as well as subgroup fairness across all protected attributes, with marked improvements on the under-performing subgroups. Beyond these quantitative results, we also conduct a comprehensive UMAP analysis that illustrates the enhanced generative performance of FairDiffusion compared to the widely-used Stable Diffusion model. To aid the analysis in this study, we also design and curate FairGenMed, the first dataset designed for assessing fairness of medical generative models, thereby promoting equitable benefits of generative AI in healthcare.

Image synthesis, or image generation, is a longstanding research problem that aims to artificially generate images that contain some particular desired content \cite{Shimoda1989New,Wandell1992The}. Prior-art techniques that tackle this problem include Variational Autoencoders (VAEs) \cite{kingma2013auto,kingma2019introduction} and Generative Adversarial Networks (GANs) \cite{goodfellow2014generative,liu2016coupled,karras2019style,zhang2019self}. VAEs suffer from posterior collapse \cite{wang2021posterior}, while GANs face challenges like mode collapse \cite{srivastava2017veegan}. In contrast, physics-inspired diffusion models have recently showed better training stability than VAEs and GANs \cite{sohl2015deep,ho2020denoising,rombach2022high}. Specifically, Rombach \etal, proposed a guiding mechanism to control image generation without retraining, and introduced cross-attention layers to enhance the model's flexibility for conditioning inputs like text or bounding boxes \cite{rombach2022high}.
Podell \etal, further enhanced the Stable Diffusion \cite{rombach2022high} model by utilizing a three times larger UNet backbone, achieved through an increase in attention blocks and a larger cross-attention context with the incorporation of a second text encoder \cite{podell2024sdxl}. However, despite the impressive performances of these diffusion models, it remains an open question how they perform across various protected attributes, particularly in sensitive domains like healthcare. Our study addresses this crucial research gap by presenting an extensive investigation into the fairness of these strong generative models.

Fairness studies in medical imaging highlight the significance of having diverse and representative datasets. To this end, various datasets have been released that advance the study of fairness within the medical domain. Specifically, in dermatology, the Fitzpatrick17k \cite{groh2021evaluating} and HAM10000 \cite{tschandl2018ham10000} datasets offer data across sensitive attributes, including various skin types, age groups, and genders. In the cardiology domain, the OL3I dataset \cite{zambrano2021opportunistic} provides heart CT data across age and gender demographics. In the context of chest x-ray analysis, the CheXpert \cite{irvin2019chexpert}, MIMIC-CXR \cite{Johnson_arXiv_2019}, COVID-CT-MD\cite{afshar2021covid} and PadChest \cite{bustos2020padchest} datasets provide chest radiographs accompanied by demographic information like age, gender, and race. The inclusion of these attributes is essential for assessing and mitigating potential biases in models trained for diagnosing chest diseases. Similarly, in the field of ophthalmology, datasets such as ODIR-2019 \cite{odir2019} and PAPILA \cite{kovalyk2022papila} provide insights into ocular conditions across various age groups and genders. Moreover, the introduction of the Harvard-GDP \cite{Luo_2023_ICCV}, GlaucomaFairness \cite{luo2023harvard}, FairSeg \cite{tian2024fairseg} and EyeFairness \cite{tian2024fairseg} datasets further advance fairness research in ophthalmology, particularly in fair glaucoma detection, by including detailed demographic attributes, encompassing age, gender, race, ethnicity, preferred language and socioeconomic factors. In particular, FairSeg \cite{tian2024fairseg} also provides additional clinical measurements such as disc and cup borders, enhancing its utility in specialized medical evaluations. While the aforementioned datasets have advanced fairness studies in medical imaging, they have not specifically addressed generative tasks. Given the growing interest in generative models, we introduce FairGenMed, an innovative dataset tailored for assessing fairness of medical generative models. Unlike existing datasets, which face challenges in fairness analysis due to noisy labels from automated extraction processes, FairGenMed provides precise ground-truth labels alongside quantitative clinical measurements such as cup-disc ratio (CDR), retinal nerve fiber layer thickness (RNFLT), and near vision refraction (NVR). Amidst the surging interest in generative models, FairGenMed has the potential to serve as a key resource in catalyzing developments in creating more equitable and unbiased generative learning algorithms.

Bayesian optimization is a strategy for efficiently optimizing black-box functions, which uses a probabilistic model, commonly instantiated as a Gaussian Process, to balance exploration and exploitation based on observed data. The method iteratively updates this model via Bayesian inference, guiding the selection of subsequent evaluation points to minimize the number of function evaluations. Bayesian optimization was first introduced by Harold Kushner~\cite{kushner1964new}, who employed Wiener processes to solve unconstrained one-dimensional optimization problems, aiming to maximize the probability of selecting samples that would lead to improvements. This approach was further developed by researchers~\cite{stuckman1988global,perttunen1990rank,elder1992global}, extending its application to high-dimensional optimization challenges and \revise{remarkably} broadening its utility. Building upon this foundational work, Mockus~\cite{movckus1975bayesian} introduced the expectation of improvement as an acquisition function. In contrast, the upper confidence bound (UCB) method balances exploration and exploitation by considering uncertainty \cite{williams2006gaussian}. In this work, we leverage Bayesian Optimization to adaptively perturb the learning process, thereby enhancing equity in medical image generation.

While our work represents the first study on the fairness of medical generative models and introduces the innovative FairDiffusion method to alleviate biases, several avenues for future work remain. Firstly, although we examine three important protected attributes -- gender, race, and ethnicity -- future studies should explore additional attributes, such as socio-economic status, to gain a more comprehensive understanding of fairness in medical generative models. Secondly, while our FairGenMed dataset is the first dataset designed for studying fairness of medical generative models, it currently only features patients from the US. Future work should collect broader datasets encompassing various geographic locations to provide a more global perspective on fairness in medical generative models. Lastly, it would be valuable to conduct studies evaluating how useful doctors find true samples versus generated samples across different demographic subgroups to gain insights into the real-world utility of these models.

To summarize, we presented the first comprehensive study investigating the fairness of medical text-to-image diffusion models. Our extensive evaluations of the widely-used Stable Diffusion model revealed \revise{notable} disparities across all protected attributes in both the image generation and semantic correlation settings. To address these fairness gaps, we proposed FairDiffusion, an equity-aware generative learning model that improves both overall performance and fairness. Moreover, we also introduced FairGenMed, the first dataset for studying fairness of medical generative models. To aid future research in fair generative learning, we make our code and dataset publicly available.

\section*{Methods}

\subsection*{FairGenMed Dataset}
We design and curate FairGenMed, the first dataset designed for assessing fairness of medical generative models. This study rigorously follows the principles delineated in the Declaration of Helsinki and has obtained approval from our institute's Institutional Review Board. All data within this dataset have been de-identified. The subjects who received glaucoma services between 2015 and 2022 were sourced from a large academic eye hospital. This study is based on scanning laser ophthalmoscopy (SLO) fundus images, which serve as a valuable marker for assessing retinal damage. The dataset includes multiple protected identity attributes, such as age, gender, race, ethnicity, preferred language, and marital status, associated with each SLO fundus image. We de-identify SLO fundus images to ensure that there is no identifying information on the images. 

Subjects are categorized into non-glaucoma and glaucoma based on their visual function assessed through visual field (VF) testing. The severity of a medical condition can be further categorized based on certain clinical measurements. For instance, the mean deviation (MD) of visual field is categorized as severe if it is less than -12 dB, moderate if it is between -12 dB and -6 dB, mild if it is between -6 dB and -1 dB, and normal otherwise. Similarly, the cup-to-disc ratio (CDR) is considered abnormal if it is greater than or equal to 0.7, borderline if it is between 0.6 and 0.7, and normal otherwise. Lastly, the near vision refraction stage is classified as negative if it is less than or equal to -0.5 diopters, neutral if it is between -0.5 and 0.5 diopters, and positive otherwise.

\begin{figure*}[t]
  \centering
    \includegraphics[width=1.\textwidth]{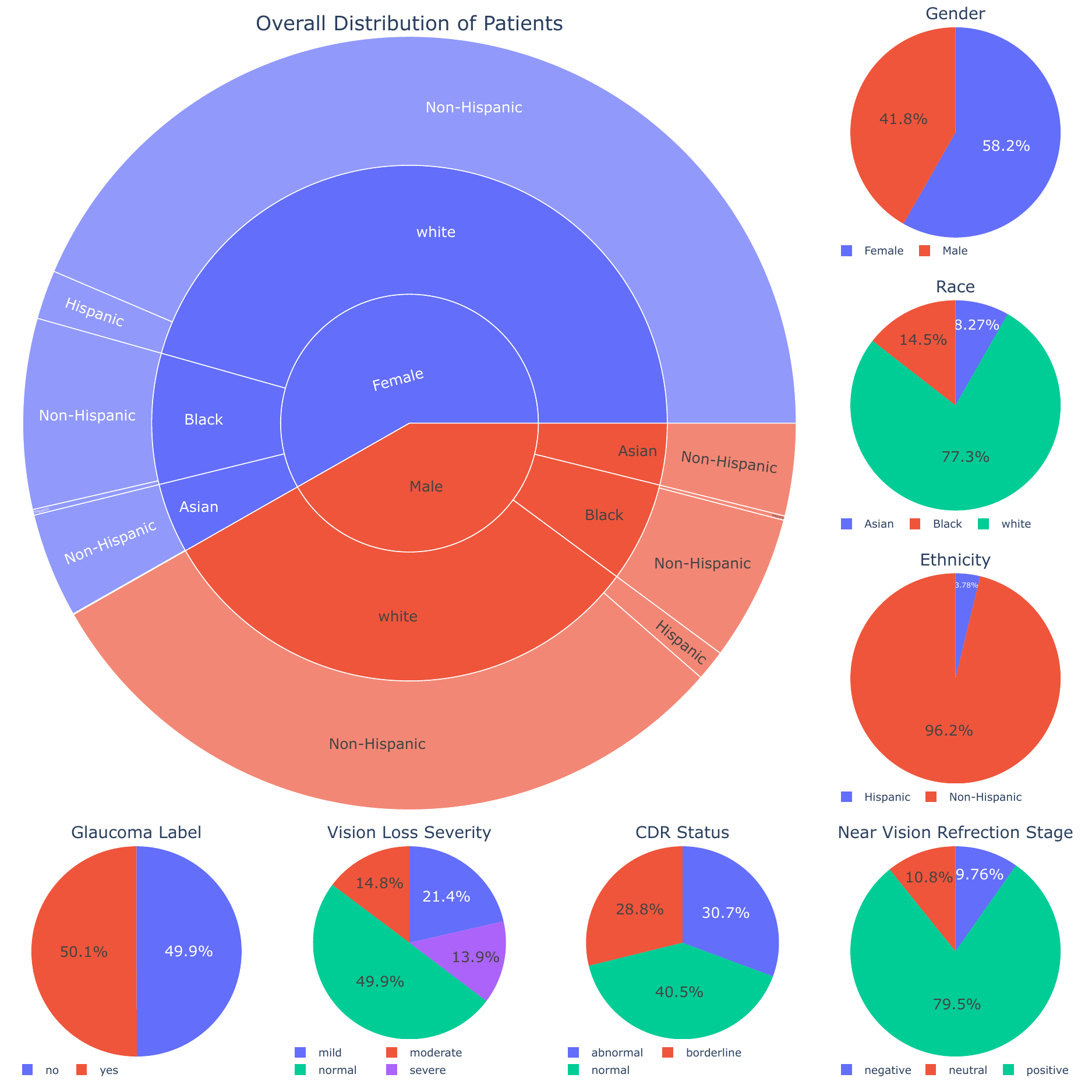}
  \caption{\revise{\textbf{Demographic and Clinical Characteristics of FairGenMed.}} We present a comprehensive overview of the patient distribution in the FairGenMed dataset, showcasing key demographic attributes and clinical features. The charts illustrate the breakdown of patients by gender, race, ethnicity, glaucoma diagnosis, MD severity, CDR status, and near vision refraction stage. This visual summary provides insights into the diversity and clinical profile of the patient population represented in the dataset.
  }
  \label{fig:stats}
\end{figure*}

The FairGenMed dataset comprises 10,000 samples from 10,000 subjects. It is divided into 6,000 training, 1,000 validation, and 3,000 test samples.
The dataset's average age is 61.6 $\pm$ 15.6 years. As shown in Fig. \ref{fig:stats}, The dataset also includes two gender categories: Female and Male, with 5,824 and 4,176 individuals, respectively. It also includes samples from three major groups: Asian, with 819 samples; Black, with  1,491 samples. In terms of ethnicity, there are two categories: Hispanic and non-Hispanic, with 9,622 and 378 individuals, respectively. 

Regarding clinical measurements, the cup-to-disc ratio is categorized as abnormal, borderline, or normal, with 3,067, 2,884, and 4,049 individuals, respectively. Moreover, the vision loss severity is categorized as mild, moderate, normal, or severe, with 2,143, 1,475, 4,994, and 1,388 individuals, respectively. Lastly, the near vision refraction stage is categorized as negative, neutral, or positive, with 926, 7,946, and 1,078 individuals, respectively. This information can be useful for analyzing the prevalence and distribution of various ocular conditions in the dataset.

\revise{
Categorical labels of cup-to-disc ratio, vision loss severity, and near vision refraction are obtained by the following formulas.\\
\textbf{Cup-to-Disc Ratio (CDR) Level:}
\[ \text{CDR} = \begin{cases} 
\text{abnormal}, & \text{if } 0.7 \leq \text{CDR} \leq 1 \\
\text{borderline}, & \text{if } 0.6 \leq \text{CDR} < 0.7 \\
\text{normal}, & \text{if } 0 \leq \text{CDR} < 0.6
\end{cases} \]
\textbf{Vision Loss Severity (VLS):}
\[ \text{VLS} = \begin{cases} 
\text{severe}, & \text{if } \text{Mean Deviation of Visual Field (MD for short)} < -12 \\
\text{moderate}, & \text{if } -12 \leq \text{MD} < -6 \\
\text{mild}, & \text{if } -6 \leq \text{MD} < -1 \\
\text{normal}, & \text{if } \text{MD} \geq -1
\end{cases} \]
where mean deviation of visual field is a measurement on a visual field test that indicates the average difference from normal expected value in the patients' particular age.\\
\textbf{Near Vision Refraction (NVR):}
\[ \text{NVR} = \text{Sphere} + 0.5 \times \text{Cylinder} \]
\[ \text{NVR Status} = \begin{cases} 
\text{negative}, & \text{if } \text{NVR} \leq -0.5 \\
\text{neutral}, & \text{if } -0.5 < \text{NVR} < 0.5 \\
\text{positive}, & \text{if } \text{NVR} \geq 0.5
\end{cases} \]
where $\text{NVR} = \text{Sphere} + 0.5 \times \text{Cylinder}$.
Sphere refers to the lens power needed to correct nearsightedness or farsightedness in an eyeglass prescription, while cylinder indicates the amount of lens power needed to correct astigmatism.
}

\subsection*{Problem Setup: Fairness in Generative Modeling}
\noindent\textbf{Stable Diffusion Model Formulation}: Stable diffusion models are a type of probabilistic models for generating new data samples that resemble the underlying distribution of training samples \cite{ho2020denoising,rombach2022high}. The mathematical formulation of a stable diffusion model involves a pair of stochastic processes: a forward diffusion process and a reverse diffusion process.

The forward diffusion process can be represented as
$q(x_t | x_0) = \mathcal{N}(x_t; \sqrt{\alpha_t}x_0, $\\
$(1 - \alpha_t)I)$, where $x_0$ is the original data sample, $x_t$ is the noisy version of the data at timestep $t$, $\alpha_t$ is a schedule of scalars that controls the amount of noise added at each timestep, and $\mathcal{N}(\cdot ; \mu, \boldsymbol{\Sigma})$ represents a Gaussian distribution with mean $\mu$ and covariance matrix $\boldsymbol{\Sigma}$. Correspondingly, the reverse diffusion process of a stable diffusion model can be represented as $p_{\theta}(x_{t-1} | x_t) = \mathcal{N}(x_{t-1}; \mu_{\theta}(x_t, t),$\\
$ \Sigma_{\theta}(x_t, t))$, where $\mu_\theta$ and $\boldsymbol{\Sigma}_\theta$ are learned functions that predict the mean and covariance of the denoised data, the parameters $\theta$ are learned by optimizing a loss function that measures the discrepancy between the forward and reverse processes. 

The model is trained to minimize the reconstruction error between the original data $x_t$ and the predicted denoised version. The loss function used for training is the denoising score matching objective, given by
$L_{L D M}:=\mathbb{E}_{\mathcal{E}(x), \epsilon \sim \mathcal{N}(0,1), t} $\\
$\left[\left\|\epsilon-\epsilon_\theta\left(z_t, t\right)\right\|_2^2\right]$, where $\mathcal{E}(x)$ is the encoding function that maps the original data $x$ to a latent space representation $z_0$, $z_t$ is the noisy version of the latent space representation at time step $t$, $\epsilon_\theta\left(z_t, t\right)$ is the neural backbone of the model, realized as a time-conditional UNet \cite{ronneberger2015u}, which predicts the noise $\epsilon$ added to the latent space representation $z_0$ at time step $t$, and $\alpha_t$ is a schedule of scalars that controls the amount of noise added at each timestep. If we further take  general conditioning inputs such as text into account, the conditional LDM objective can be rewritten as 
\begin{align}
    L_{L D M}:=\mathbb{E}_{\mathcal{E}(x), y, \epsilon \sim \mathcal{N}(0,1), t}\left[\left\|\epsilon-\epsilon_\theta\left(z_t, t, \tau_\theta(y)\right)\right\|_2^2\right]
    \label{eqn:ldm_obj}
\end{align}
where $\tau_\theta$ is a text encoder.

\noindent\textbf{Demographic Discrepancy in Generative Modeling}: The discrepancy between different demographic groups in terms of image generation performance can be formulated as a problem of distribution mismatch. The misalignment would result in biased AI models.

Let's denote $P(x|\{\mathcal{A}^{i}|1\le i \le 3\})$ as the distribution of images $x$ conditioned on demographic attributes $\{\mathcal{A}^{i}\}$. Specifically, there are three types of attributes in this work, \ie, race ($i=1$) $\mathcal{A}^{1}\in \{ \text{Asian}, \text{Black}, \text{\revise{white}} \}$, gender ($i=2$) $\mathcal{A}^{2}\in \{ \text{Female}, \text{Male} \}$, and ethnicity ($i=3$) $\mathcal{A}^{3}\in \{ \text{Non-Hispanic}, \text{Hispanic} \}$. The goal is to learn a generative model G that can sample images $x \sim G(x|\{\mathcal{A}^{i}\})$ and aims to ensure that all demographic groups should be well-represented.

The discrepancy between the generated images and the real images for a specific demographic group d can be measured using a suitable distance metric D, such as the Frechet Inception Distance (FID) \cite{heusel2017gans} (or the Inception Score (IS) \cite{salimans2016improved}). The objective of training generative models can then be formulated as
\begin{align}
    \begin{split}
        \min_{G} & \text{ FID}(P(x|\{\mathcal{A}^{i}\}), G(x|\{\mathcal{A}^{i}\})) \\
        \text{s.t.} & \text{ FID}(P(x|\mathcal{A}^{1}_{j},\{\mathcal{A}^{i\ne 1}\})), G(x|\mathcal{A}^{1}_{k},\{\mathcal{A}^{i\ne 1}\})) \leq \epsilon, \ \ j \ne k
    \end{split}
    \label{eqn:obj}
\end{align}
where $\epsilon$ is a pre-specified tolerance level for the discrepancy between the generated images and the real images for each demographic group. The objective (\ref{eqn:obj}) not only optimizes the FID between two distributions but also aims to minimize the discrepancy between the distributions with respect to any two racial groups while assuming the other demographic attributes are the same. Smaller $\epsilon$ implies the model G performs more fairly for the demographic groups. Specifically, the objective (\ref{eqn:obj}) focuses on the race attribute for simplicity, but it can be generalized to other attributes as well.

\subsection*{FairDiffusion}
The core idea of the proposed FairDiffusion (FD) method is to adaptively perturb the learning process based on demographic attributes in order to achieve fairness in latent diffusion models. Specifically, the method determines how to perturb the learning process with samples from each demographic group. FairDiffusion employs a Bayesian Optimization based approach, named Fair Bayesian Perturbation, to find the optimal mapping between the perturbations applied to each demographic group and the resulting quantifiable fairness.

\noindent\textbf{Fair Bayesian Perturbation}: We hypothesize that there is a correlation between the perturbations applied to each demographic group and the resulting quantifiable fairness. Intuitively, alleviating the perturbation to none degenerates the proposed method to the standard procedure. Given samples $\{x, y, \{\mathcal{A}^{i}\} \}$ ($1\le i\le 3$), based on the LDM objective (\ref{eqn:ldm_obj}), the perturbation on the learning process of FairDiffusion can be formulated as
\begin{align}
\begin{split}
    L_{FD}:=&\mathbb{E}_{\mathcal{E}(x), y, \{\mathcal{A}^{i}\}, \epsilon \sim \mathcal{N}(0,1), t}\left[ \left(1 + \zeta_{\{\mathcal{A}^{i}\}} \right) \left\|\epsilon-\epsilon_\theta\left(z_t, t, \tau_\theta(y)\right)\right\|_2^2\right] \\
    \text{s.t. } &\zeta_{\{\mathcal{A}^{i}\}} = \sum_{i}\sum_{j} \zeta_{\mathcal{A}^{i}_{j}}, \ \  \zeta_{\mathcal{A}^{i}_{j}} \sim \mathcal{N}(0, \psi_{\mathcal{A}^{i}_{j}}),
\end{split}
    \label{eqn:fd_obj}
\end{align}
where $\zeta_{\{\mathcal{A}^{i}\}}$ is the conditioned perturbation sampled from various distributions that are associated with attributes $\mathcal{A}^{i}$. $\zeta_{\{\mathcal{A}^{i}\}} \to 0$ implies $L_{FD}$ degenerates to $L_{LDM}$. In the meantime, the fairness can be represented by the discrepancy between the demographic group achieving the largest average loss and the one achieving the lowest average loss. It can be mathematically formulated as
\begin{align}
\begin{split}
    \Delta &= \sum_{i=1}^{|\mathcal{A}|} \left(\frac{1}{N_{ \mathcal{A}^{i}_{max}}} \sum_{(x, y| \mathcal{A}^{i}_{max}) \in \mathcal{B}} \ell(x, y) - \frac{1}{N_{ \mathcal{A}^{i}_{min}}} \sum_{(x', y'| \mathcal{A}^{i}_{min}) \in \mathcal{B}} \ell(x', y') \right ) \\
    \text{s.t. }&  \mathcal{A}^{i}_{max} = \argmax_{\mathcal{A}^{i}} \frac{1}{N_{ \mathcal{A}^{i}}} \sum_{(x, y| \mathcal{A}^{i}) \in \mathcal{B}} \ell(x, y), \ \  \mathcal{A}^{i}_{min} = \argmin_{\mathcal{A}^{i}} \frac{1}{N_{ \mathcal{A}^{i}}} \sum_{(x, y| \mathcal{A}^{i}) \in \mathcal{B}} \ell(x, y),
\end{split}
    \label{eqn:gap_loss}
\end{align}
where $\mathcal{B}$ is a batch, $N_{ \mathcal{A}^{i}}$ is the number of samples with attribute $\mathcal{A}^{i}$, and $\ell(x, y) = \left\|\epsilon-\epsilon_\theta\left(z_t, t, \tau_\theta(y)\right)\right\|_2^2$ is an instance-level loss.

To effectively enhance fairness through adaptive perturbation in the learning process, it is crucial to understand the correlation between the perturbation generated by Equation (\ref{eqn:fd_obj}) and the group-wise loss discrepancy $\Delta$. Essentially, the proposed method aims to learn the mapping between attribute-specific Gaussian distribution parameters $\Psi = \{\psi_{\mathcal{A}^{i}_{j}}\in \mathbb{R}| i\in\{1,2,3\}, j\in\{1,2,3\}\  (\text{or } j\in\{1,2\}) \}$ and the group-wise loss discrepancy $\Delta \in \mathbb{R}$. Note that $j\in \{1,2\}$ when $i=2,3$.

With observations $D=\{\Psi_{t}, -\Delta_{t}\}_{t=t_{0}}^{t_{0}+w}$ collected from the past iterations within a pre-defined time window $w$, we can formulate a Bayesian Optimization problem to optimize an acquisition function evaluated on a Gaussian posterior $f_{\mathcal{D}}(\Psi)=\mathcal{N}\left(\mu_\Psi, \sigma_\Psi^2\right)$ over the candidate set $\Psi$. General acquisition functions can be written as
\begin{align}
a\left( \Psi ; \Phi, D \right) =\mathbb{E}\left[a\left(g\left(f \left(\Psi \right) \right), \Phi\right) \mid D\right]
\end{align}
where $g$ is an objective function, $\Phi \in \boldsymbol{\Phi}$ are parameters independent of $\Psi$ in the set of parameters $\boldsymbol{\Phi}$, and $a: \mathbb{R}^q \times \mathbf{\Phi} \rightarrow \mathbb{R}$ is a utility function that defines the acquisition function. We adopt the upper confidence bound (UCB) \cite{williams2006gaussian} method as the acquisition function. Then, the optimal $\Psi$ can be determined by
\begin{align}
    \Psi^{*} = \argmax_{\Psi} a(\Psi ; \Phi, D)
    \label{eqn:bo_solution}
\end{align}
Note that $\Delta$ is negated to align with the paradigm of Bayesian Optimization, where the goal is to maximize the acquisition function to find the optimal solution.

\noindent\textbf{Knowledge Exploration-Exploitation Mechanism}: Knowledge exploration-exploitation is vital in machine learning as it maintains a delicate balance between discovering new information and leveraging existing knowledge to enhance learning outcomes. The optimal solution, computed by Equation (\ref{eqn:bo_solution}), relies on observations $\{\Psi_{t}, -\Delta_{t}\}$, emphasizing the importance of balancing exploration and exploitation. To achieve this balance, an exploitation rate $\nu$ is introduced, with $1-\nu$ representing the exploration rate. The update of $\Psi$ at the $(t+1)$-th iteration is then given by:
\begin{align}
\Psi_{t+1} \leftarrow \nu\Psi_{t} + (1-\nu) \hat{\zeta}, \ \ \text{where } \ \hat{\zeta} \sim \mathcal{N}(0,1)
\end{align}

\subsection*{Training Diffusion Models}
Next, we briefly describe the setup for training the diffusion models. We train a Stable Diffusion~\cite{rombach2022high} model on SLO fundus images following the widely-used huggingface implementation~\cite{von-platen-etal-2022-diffusers}. Specifically, we initialize from the official Stable Diffusion~\cite{rombach2022high} checkpoints and fine-tune on 7,000 SLO fundus images from the proposed FairGenMed dataset. Following the huggingface implementation~\cite{von-platen-etal-2022-diffusers}, we only fine-tune the UNet while keeping the text encoder and VAE frozen, and use the following prompt for conditioning the diffusion model on various demographic and clinical attributes: \textit{SLO fundus image of a [race, gender, ethnicity] patient with the following conditions: [glaucoma, cup-disc ratio, retinal nerve fiber layer thickness status, near vision refraction status]}. We employ a learning rate of 1e-4, AdamW optimizer with $\beta_{1}=0.9$, $\beta_{2}=0.99$, $wd=1e-2$, and train on 2 A100 GPUs with a per-device batch size of 16. For training the FairDiffusion model, we employ $w=30$ and $\nu=0.95$. Once the diffusion model is trained, we conduct two types of quantitative evaluations, including both generation and classification metrics.

\subsection*{Evaluation Metrics}
Lastly, we describe two sets of metrics (generation and classification) to assess the fairness across image generation as well as semantic correlation.

\noindent\textbf{Generation Metrics}:
In safety-critical medical applications, relying solely on either fairness or accuracy as the sole measurement criterion is insufficient \cite{luo2023harvard,tian2024fairseg}. Therefore, we follow \cite{luo2023harvard,tian2024fairseg} to use equity-scaled metrics to measure the performance for both the classification and generation tasks. In the task of image generation, widely-used metrics FID \cite{heusel2017gans} and the Inception Score IS \cite{salimans2016improved} can be extended to the equity-scaled versions, enabling a nuanced understanding of the trade-off between fairness and accuracy. ES-FID and ES-IS are defined as:
{\small
\begin{equation*}
    \begin{split}
    \text{ES-FID}_{\mathcal{A}^{i}} = \left( 1+\frac{\frac{1}{|\mathcal{A}^{i}|}\sum_{j=1}^{|\mathcal{A}^{i}|}| \text{FID} - \text{FID}_{\mathcal{A}^{i}_{j}} |}{\text{FID}} \right) \text{FID}, \quad \text{ES-IS}_{\mathcal{A}^{i}} = \frac{\text{IS}}{1+\sum_{j=1}^{|\mathcal{A}^{i}|}| \text{IS} - \text{IS}_{\mathcal{A}^{i}_{j}} |}
\end{split}
\label{eqn:es_metric}
\end{equation*}
}
where $\text{ES-IS}_{\mathcal{A}^{i}}$ means ES-IS on attribute $\mathcal{A}^{i}$, such as race, gender, or ethnicity. Note that lower FID scores indicate better performance, whereas higher IS scores signify superior performance. Thus, the form of ES-FID is different from that of ES-IS, and they have distinct formulations. Nonetheless, both metrics share the concept that a smaller discrepancy (\eg, $\sum_{j=1}^{|\mathcal{A}^{i}|}| \text{IS} - \text{IS}_{\mathcal{A}^{i}_{j}} |$) leads to a better trade-off between fairness and accuracy. For instance, ES-IS tends to be IS as $\sum_{j=1}^{|\mathcal{A}^{i}|}| \text{IS} - \text{IS}_{\mathcal{A}^{i}_{j}} | \to 0$. We evaluate generation quality of the trained diffusion models using FID, ES-FID, IS, and ES-IS with respect to the test set of 3000 images from FairGenMed. In addition to the overall generation performance, we also report the generation metrics on individual subgroups within the race, gender, and ethnicity attributes in order to understand whether the quality of generated images is consistent across the different demographic subgroups.

\revise{In addition, we adopt a widely-adopted and straightforward fairness metric maximum discrepancy (MD) of FID or IS on various demographic attributes \cite{lahoti2020fairness,dullerud2022is}. Specifically, given FID or IS scores on demographic subgroups, maximum discrepancy indicates the gap between the highest score and lowest score. Lower MD values indicate better equitable generation performance.
\\
It is noteworthy that MD of FID or IS only focuses on equity, which may not not fully capture the clinical relevance and accuracy required in safety-critical medical applications. For example, consider two different models generating fundus images for glaucoma diagnosis across three demographic groups (Asian, Black, and \revise{white}). Model A achieves FID scores of 95, 98, and 96 respectively across these groups, resulting in an MD of only 3. Model B achieves FID scores of 45, 55, and 35, leading to an MD of 20. While model A would appear more "equitable" by the MD metric alone due to its smaller maximum discrepancy, it produces uniformly poor quality images across all groups (FID scores above 95). In contrast, model B, despite having a larger MD, generates much higher quality images (FID scores between 35-55) that would be more clinically useful. This underscores the importance of considering the trade-off between equity and model performance when evaluating fairness in the context of medical applications.
}

\noindent\textbf{Classification Metrics}: Besides generation metrics, we also conduct a classification task to study the semantic correlation between text prompts and the generated SLO fundus images. Specifically, we construct a generated dataset by sampling from the trained diffusion model following the same distribution as the training set of FairGenMed. Next, we train a classification model on this generated dataset and evaluate it on the original test set from FairGenMed. In addition to the overall classification performance measured via AUC, we also report the fairness metrics Equity-Scaled AUC (ES-AUC) and Difference in Equalized Odds (DEOdds)~\cite{hardt2016equality}, as well as the individual group-wise AUCs. We use two classification models -- ViT-B and EfficientNet. ViT-B is trained following the official MAE~\cite{he2022masked} settings with a base learning rate of 5e-4, weight decay of 0.01, layer decay of 0.55, drop path rate of 0.1, and batch size of 64 for 50 epochs. EfficientNet is trained with a learning rate of 1e-3 and AdamW optimizer for 10 epochs. Both models are trained on a single A100 GPU.

\clearpage 

%
\bibliography{ref2} 
\bibliographystyle{sciencemag}

%
%
%
%
%
%


\section*{Acknowledgments}
This work was supported by NIH R00 EY028631, NIH R01 EY036222, Research to Prevent Blindness International Research Collaborators Award, Alcon Young Investigator Grant, and NIH P30 EY003790. We also acknowledge the generous funding resources provided by NYU Abu Dhabi with code AD131.

\paragraph*{Funding:}
Please refer to the section of Acknowledgements.
\paragraph*{Author contributions:}
The concept for the study was developed by Yan Luo, Muhammad Osama Khan, and Mengyu Wang. Data acquisition, analysis, and interpretation were done by Yan Luo, Muhammad Osama Khan, Congcong Wen, and Muhammad Muneeb Afzal. Methodological design and implementation were done by Yan Luo and Muhammad Osama Khan. Congcong Wen verified the effectiveness of the proposed method on external datasets CheXpert and HAM10000. Figure 1 was created by Titus Fidelis Wuermeling. The code review was carried out by Congcong Wen, Min Shi, and Yu Tian. The paper was written by Yan Luo and Muhammad Osama Khan. Critical revision of the paper was carried out by all the authors. The study was supervised by Yi Fang and Mengyu Wang.

\paragraph*{Competing interests:}
There are no competing interests to declare.

\paragraph*{Data and materials availability:}

All data needed to evaluate the conclusions of the paper are present in the paper and the Supplementary Materials.
The dataset and code have been uploaded to Zenodo. The dataset is publicly accessible via the DOI link \url{https://doi.org/10.5281/zenodo.13178701}, while the code is publicly accessible via the DOI link \url{https://doi.org/10.5281/zenodo.14606588}.





\newpage


\renewcommand{\thefigure}{S\arabic{figure}}
\renewcommand{\thetable}{S\arabic{table}}
\renewcommand{\theequation}{S\arabic{equation}}
\renewcommand{\thepage}{S\arabic{page}}
\setcounter{figure}{0}
\setcounter{table}{0}
\setcounter{equation}{0}
\setcounter{page}{1} 


\begin{center}
\section*{Supplementary Materials for\\ \scititle}

Yan Luo$^{1,2,6\dagger}$,
	Muhammad Osama Khan$^{4\dagger}$,
        Congcong Wen$^{3,4\dagger}$, \\
	Muhammad Muneeb Afzal$^{4}$, 
     Titus Fidelis Wuermeling$^{1,2}$,
     Min Shi$^{1,2}$,
     Yu Tian$^{1,2}$,\\
     Yi Fang$^{3,4\mathsection}$,
     Mengyu Wang$^{1,2,5,6\ast\mathsection}$\\ 
\small$^\ast$Corresponding author: Mengyu Wang. Email: mengyu\_wang@meei.harvard.edu\\
\small$^\dagger$These authors contributed equally as co-first authors. 
    \small$^\mathsection$These authors contributed equally.
\end{center}

\subsubsection*{This PDF file includes:}
Training Scheme\\
Generation Metrics\\
Classification Metrics\\
UMAP analysis of Gender, Race, and Ethnicity\\
Ablation Studies\\
Figures S1 to S5\\
Computational Complexity Analysis\\


\newpage


\section*{Training Scheme}
We train a Stable Diffusion~\cite{rombach2022high} model on SLO fundus images following the widely-used huggingface implementation~\cite{von-platen-etal-2022-diffusers}. Specifically, we initialize from the official Stable Diffusion~\cite{rombach2022high} checkpoints and fine-tune on 6000 SLO fundus images from the proposed FairGenMed dataset. Following the huggingface implementation~\cite{von-platen-etal-2022-diffusers}, we only fine-tune the UNet while keeping the text encoder and VAE frozen, and use the following prompt for conditioning the diffusion model on various demographic and clinical attributes: \textit{SLO fundus image of a [race, gender, ethnicity] patient with the following conditions: [glaucoma, cup-disc ratio, retinal nerve fiber layer thickness status, near vision refraction status]}. We employ a learning rate of 1e-4, AdamW optimizer with $\beta_{1}=0.9$, $\beta_{2}=0.99$, $wd=1e-2$, and train on 2 A100 GPUs with a per-device batch size of 16. For training the FairDiffusion model, we employ $w=30$ and $\nu=0.95$. Once the diffusion model is trained, we conduct two types of quantitative evaluations, including both generation and classification metrics.

\begin{algorithm}
\caption{Fair Bayesian Perturbation}\label{alg:proposed}
\begin{algorithmic}[1]
\Require $\{x, y, \{\mathcal{A}^{i}\} \}$ ($1\le i\le 3$), $\epsilon_{\theta}$, $\epsilon_{\theta}$, $\tau_\theta$, $w$, $\nu$
\State Initialize $\Psi$
\While{ not reach the pre-defined number of iterations }
\State Compute FairDiffusion loss using Equation (\ref{eqn:fd_obj})
\State Compute group-wise loss discrepancy $\Delta$ using Equation (\ref{eqn:gap_loss})
\State Put the observation $(\Psi_{t}, -\Delta_{t})$ to the stack $D$
\State Optimize acquisition function $a(\Psi ; \Phi, D)$ using Bayesian Optimization
\If{$|D|=w$}
    \State Update $\Psi \leftarrow \argmax_{\Psi} a(\Psi ; \Phi, D)$ using Equation (\ref{eqn:bo_solution})
    \State Empty $D$
  \EndIf
\State Update $\Psi \leftarrow \nu\Psi + (1-\nu) \hat{\zeta}$ with exploitation rate $\nu$
\EndWhile
\State \textbf{return} $\Psi^{}$
\end{algorithmic}
\end{algorithm}

\section*{Generation Metrics}
In safety-critical medical applications, relying solely on either fairness or accuracy as the sole measurement criterion is insufficient \cite{luo2023harvard,tian2024fairseg}. Therefore, we follow \cite{luo2023harvard,tian2024fairseg} to use equity-scaled metrics to measure the performance for both the classification and generation tasks. In the task of image generation, widely-used metrics FID \cite{heusel2017gans} and the Inception Score IS \cite{salimans2016improved} can be extended to the equity-scaled versions, enabling a nuanced understanding of the trade-off between fairness and accuracy. ES-FID and ES-IS are defined as:
{\small
\begin{equation*}
    \begin{split}
    \text{ES-FID}_{\mathcal{A}^{i}} = \left( 1+\frac{\frac{1}{|\mathcal{A}^{i}|}\sum_{j=1}^{|\mathcal{A}^{i}|}| \text{FID} - \text{FID}_{\mathcal{A}^{i}_{j}} |}{\text{FID}} \right) \text{FID}, \ \text{ES-IS}_{\mathcal{A}^{i}} = \frac{\text{IS}}{1+\sum_{j=1}^{|\mathcal{A}^{i}|}| \text{IS} - \text{IS}_{\mathcal{A}^{i}_{j}} |}
\end{split}
\label{eqn:es_metric}
\end{equation*}
}
where $\text{ES-IS}_{\mathcal{A}^{i}}$ means ES-IS on attribute $\mathcal{A}^{i}$, such as race, gender, or ethnicity. Note that lower FID scores indicate better performance, whereas higher IS scores signify superior performance. Thus, the form of ES-FID is different from that of ES-IS, and they have distinct formulations. Nonetheless, both metrics share the concept that a smaller discrepancy (\eg, $\sum_{j=1}^{|\mathcal{A}^{i}|}| \text{IS} - \text{IS}_{\mathcal{A}^{i}_{j}} |$) leads to a better trade-off between fairness and accuracy. For instance, ES-IS tends to be IS as $\sum_{j=1}^{|\mathcal{A}^{i}|}| \text{IS} - \text{IS}_{\mathcal{A}^{i}_{j}} | \to 0$. We evaluate generation quality of the trained diffusion models using FID, ES-FID, IS, and ES-IS with respect to the test set of 3000 images from FairGenMed. In addition to the overall generation performance, we also report the generation metrics on individual subgroups within the race, gender, and ethnicity attributes in order to understand whether the quality of generated images is consistent across the different demographic subgroups.


\begin{figure}[t]
  \centering
    \includegraphics[width=0.8\textwidth]{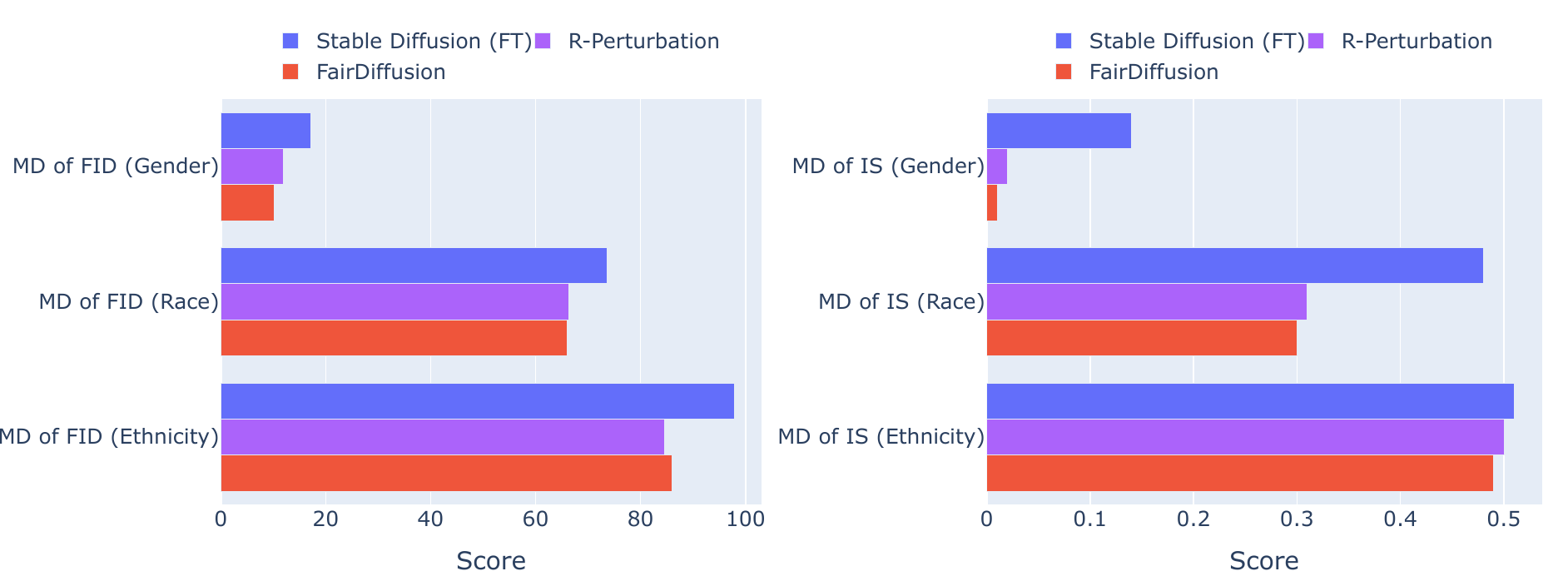}
  \caption{\revise{\textbf{Maximum discrepancy (MD) analysis of generation quality across demographic attributes.}} Left: Maximum discrepancy of FID scores across gender, race, and ethnicity subgroups. Lower MD of FID indicates more consistent generation quality across demographic subgroups. Right: Maximum discrepancy of IS scores across demographic attributes. Lower MD of IS similarly suggests more equitable generation performance. FairDiffusion achieves lower maximum discrepancy compared to both Stable Diffusion (FT) and R-Perturbation baselines, indicating more consistent generation quality across different demographic groups.}
  \label{fig:md_fid_is}
\end{figure}

\section*{Classification Metrics}
Besides generation metrics, we also conduct a classification task to study the semantic correlation between text prompts and the generated SLO fundus images. Specifically, we construct a generated dataset by sampling from the trained diffusion model following the same distribution as the training set of FairGenMed. Next, we train a classification model on this generated dataset and evaluate it on the original test set from FairGenMed. In addition to the overall classification performance measured via AUC, we also report the fairness metrics Equity-Scaled AUC (ES-AUC) and Difference in Equalized Odds (DEOdds)~\cite{hardt2016equality}, as well as the individual group-wise AUCs. We use two classification models -- ViT-B and EfficientNet. ViT-B is trained following the official MAE~\cite{he2022masked} settings with a base learning rate of 5e-4, weight decay of 0.01, layer decay of 0.55, drop path rate of 0.1, and batch size of 64 for 50 epochs. EfficientNet is trained with a learning rate of 1e-3 and AdamW optimizer for 10 epochs. Both models are trained on a single A100 GPU.

\section*{UMAP analysis of Gender, Race, and Ethnicity}
\begin{figure*}[t]
  \centering
    \includegraphics[width=1.\textwidth]{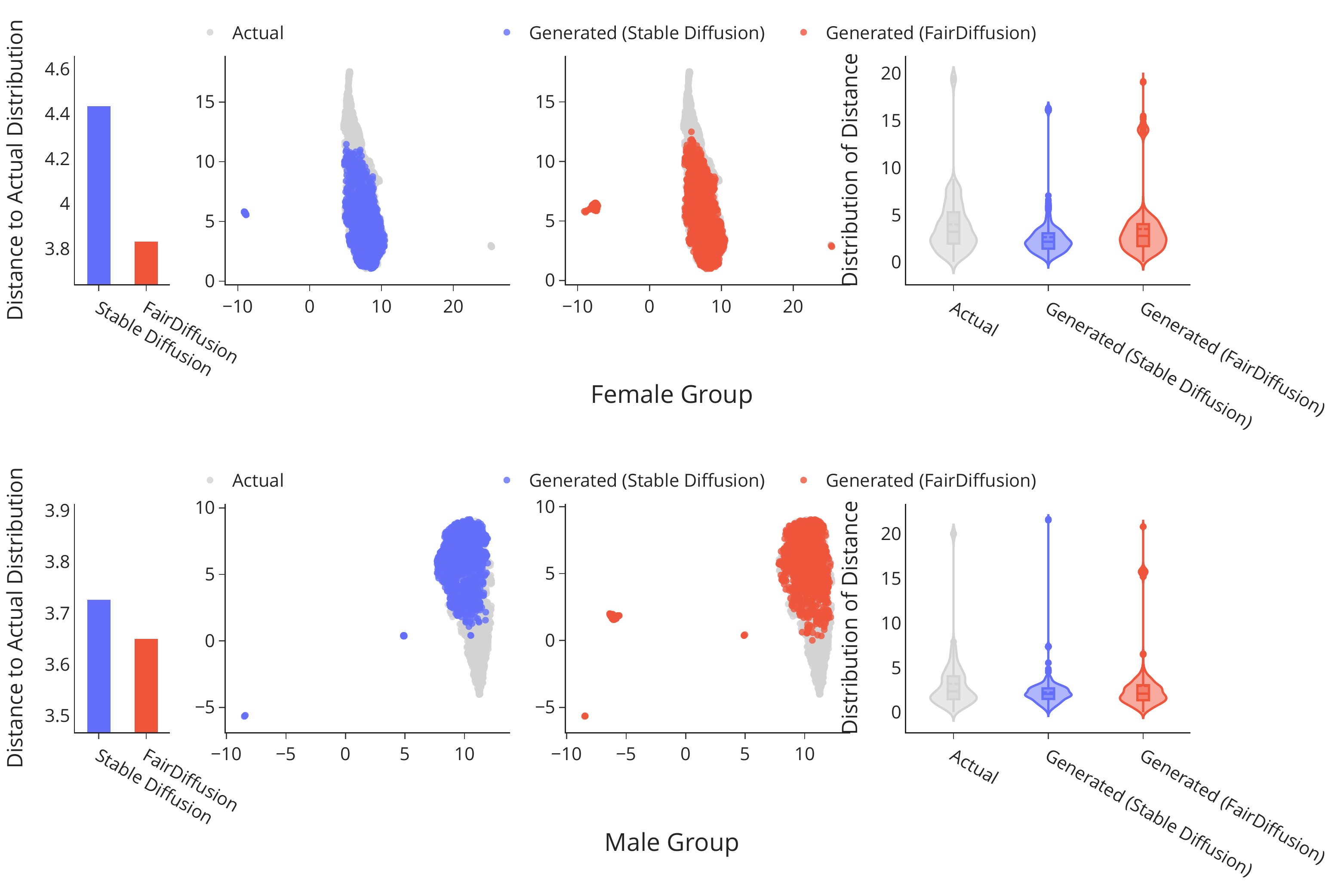}
  \caption{\revise{\textbf{UMAP Analysis of Image Generation Models on Gender.}}
  }
  \label{fig:gender_umap}
\end{figure*}

\begin{figure*}[t]
  \centering
    \includegraphics[width=1.\textwidth]{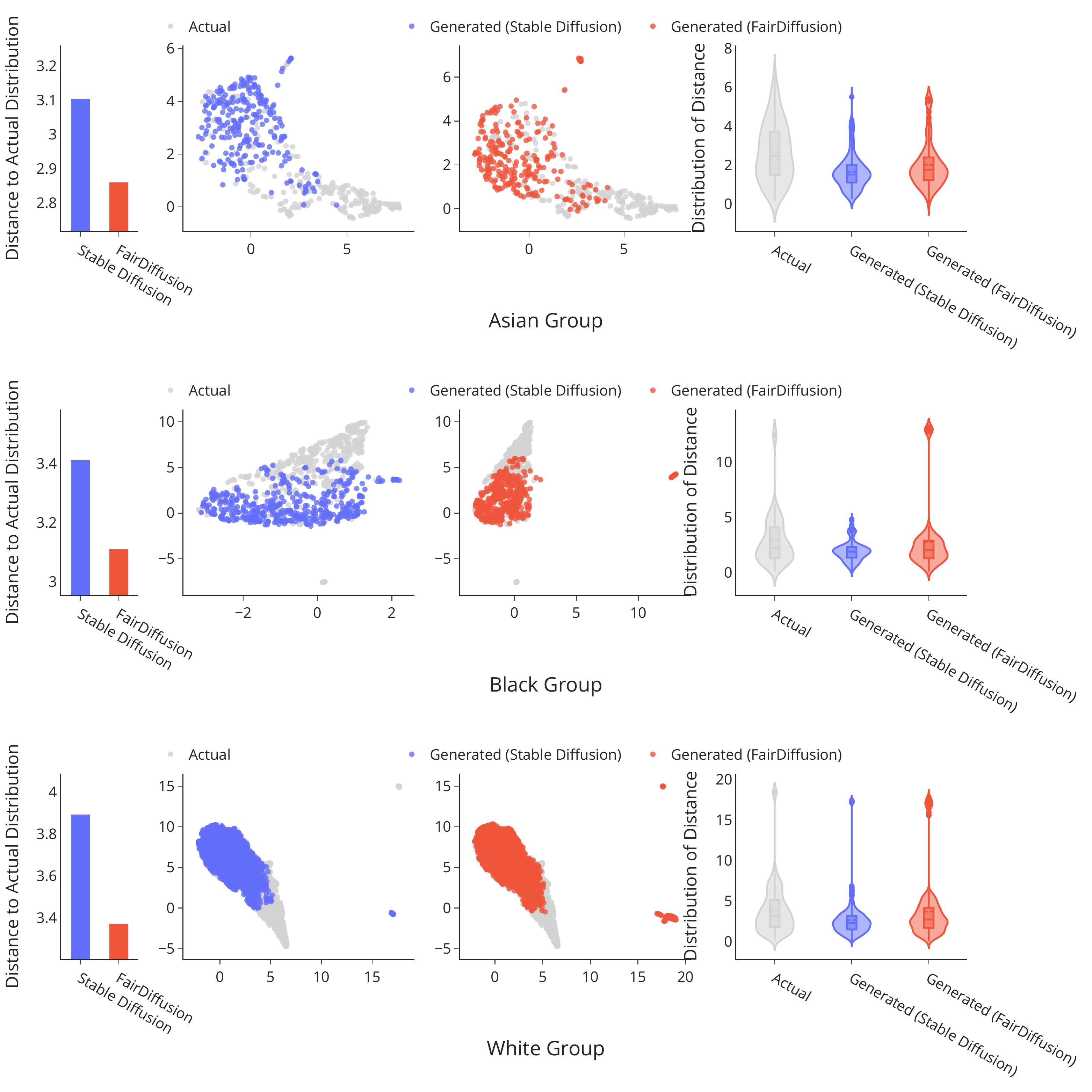}
  \caption{\revise{\textbf{UMAP Analysis of Image Generation Models on Race.}}
  }
  \label{fig:race_umap}
\end{figure*}

\begin{figure*}[t]
  \centering
    \includegraphics[width=1.\textwidth]{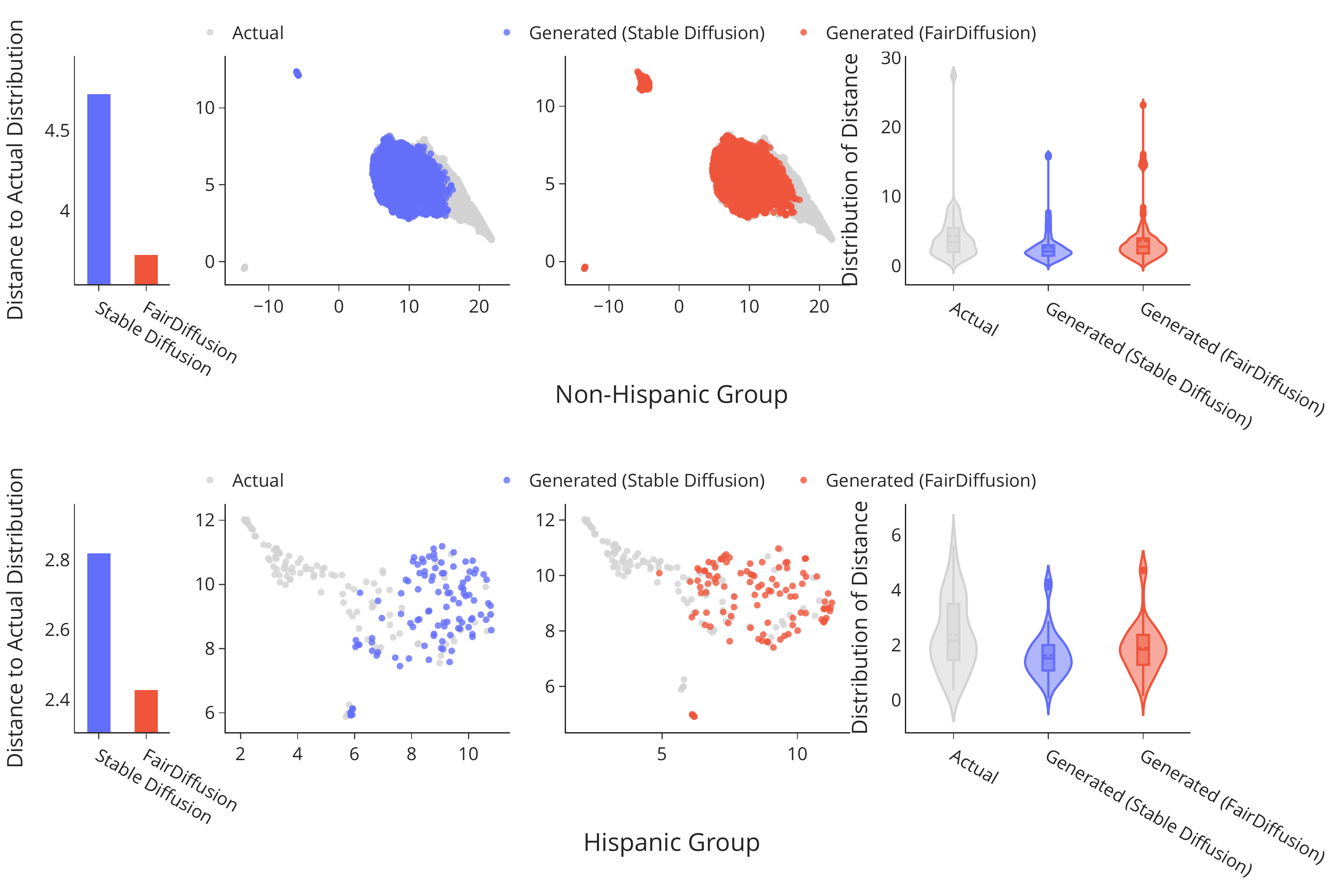}
  \caption{\revise{\textbf{UMAP Analysis of Image Generation Models on Ethnicity.}}
  }
  \label{fig:ethnicity_umap}
\end{figure*}

Fig. \ref{fig:gender_umap}, \ref{fig:race_umap}, \ref{fig:ethnicity_umap} show UMAP analyses of image generation models on Gender, Race, and Ethnicity, respectively. 

\section*{Ablation Studies}
\begin{figure}[t]
\begin{minipage}{.48\linewidth}
\centering
    \includegraphics[width=\textwidth]{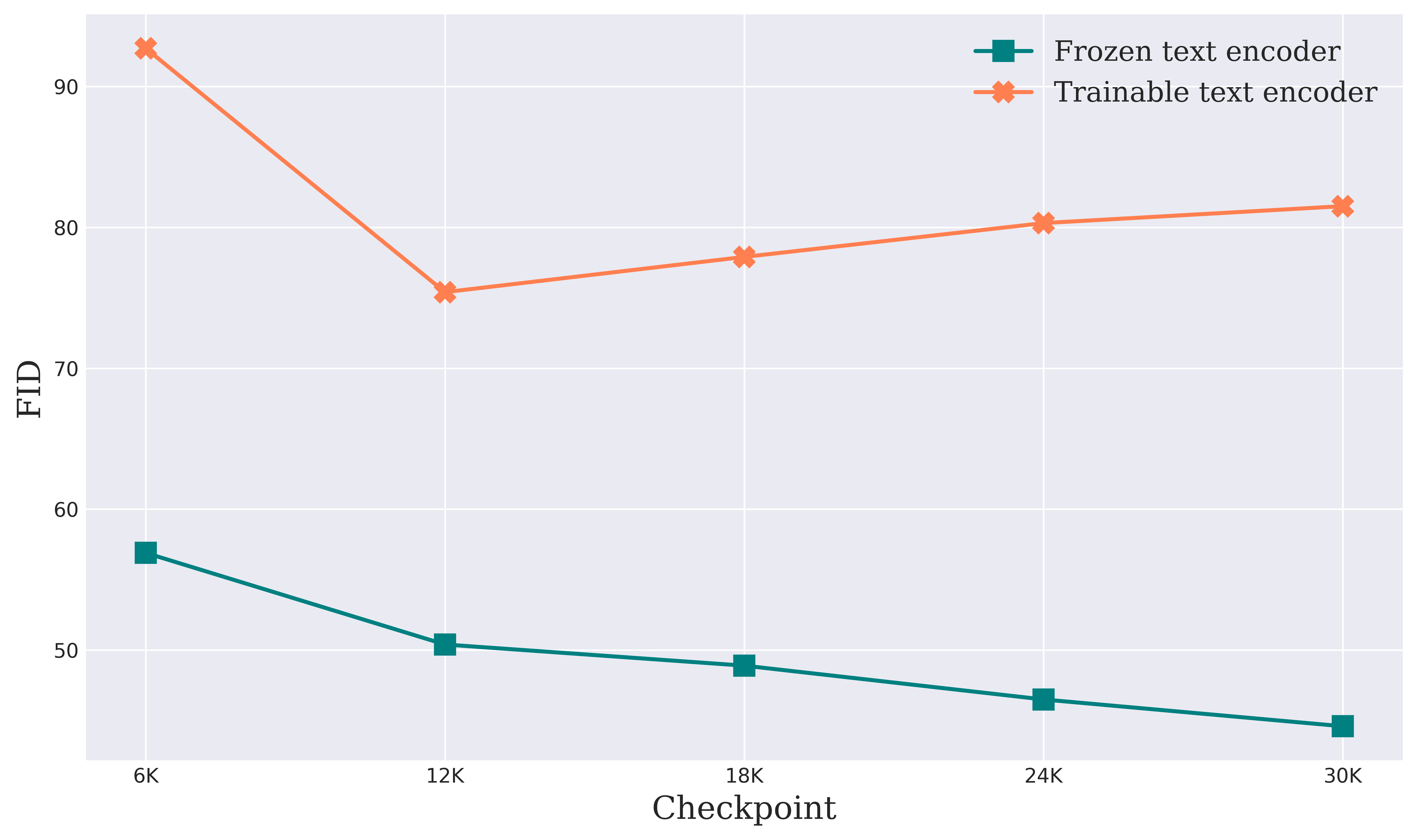}
\end{minipage}\hfill
\begin{minipage}{.5\linewidth}
    \includegraphics[width=\textwidth]{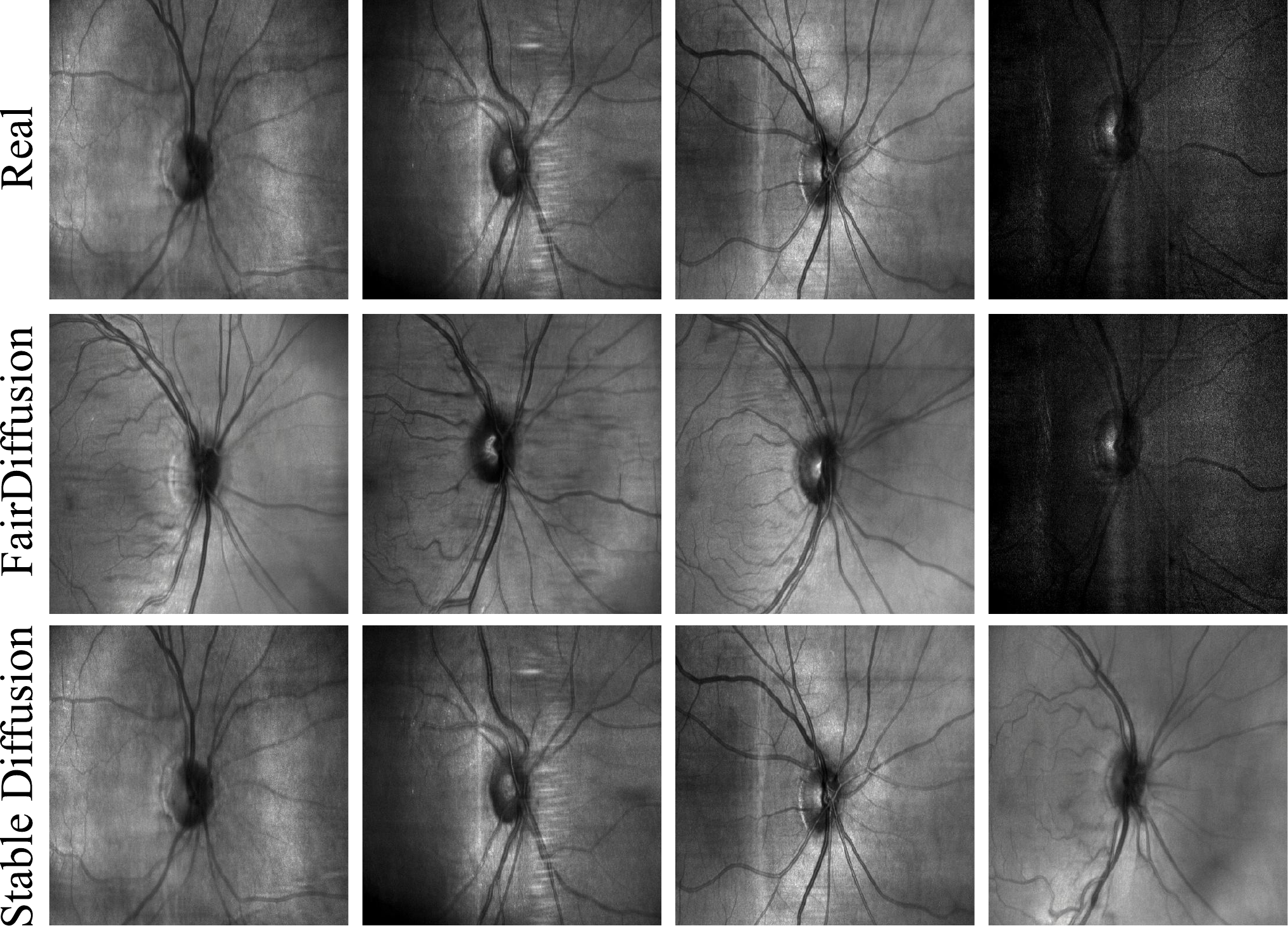}
\centering
\end{minipage}
\caption{\revise{\textbf{Ablation studies of Stable Diffusion (SD) and FairDiffusion.}} \textbf{Left: }Frozen vs. trainable text encoder for SD v2.1. \textbf{Right: }Qualitative examples generated by the Stable Diffusion model and the proposed FairDiffusion. Images in each column have the same combination of attributes and they are generated under the same settings.}
\label{fig:ablations_sd}
\end{figure}

In this section*, we conduct detailed ablation studies to rigorously investigate the impact of different components on model performance and fairness. Firstly, Figure~\ref{fig:ablations_sd} (left) explores the performance of frozen versus trainable text encoders, showing that the frozen text encoder performs better, which is consistent with the widely-used huggingface implementation~\cite{von-platen-etal-2022-diffusers}. Secondly, Figure~\ref{fig:ablations_sd} (right) illustrates several qualitative examples indicating the superiority of the FairDiffusion generations. The supplementary material presents further ablation studies on investigating the effectiveness of EMA (exponential moving average) versus non-EMA Stable Diffusion UNet. Moreover, we also conduct several investigations to verify the effectiveness of different design choices of the proposed FairDiffusion method.

\begin{figure}[t]
\centering
\includegraphics[width=0.8\textwidth]{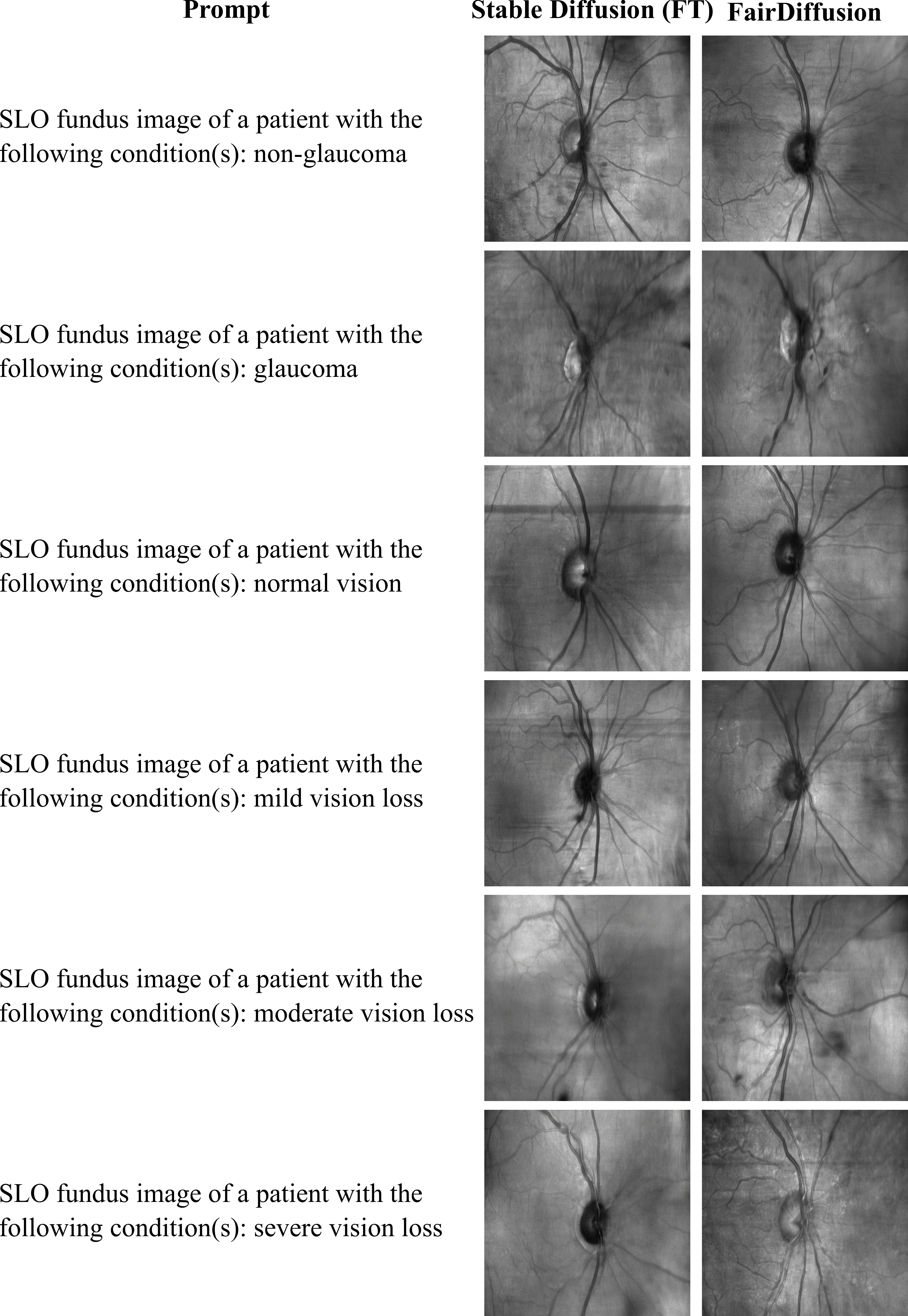}
\caption{\revise{\textbf{Comparison of fundus images generated by Stable Diffusion (FT) and FairDiffusion models for various ophthalmic conditions.}} The image-prompt pairs show SLO fundus images for six different conditions: non-glaucoma, glaucoma, normal vision, mild vision loss, moderate vision loss, and severe vision loss. }
\label{fig:more_example}
\end{figure}

\section*{Computational Complexity Analysis} 

\revise{\revise{\textbf{Our analysis demonstrates that FairDiffusion achieves improved performance and fairness without significant computational overhead compared to Stable Diffusion.}} As shown in Table~\ref{tab:training_time}, the per-sample training times of FairDiffusion and Stable Diffusion are comparable. This efficiency stems from FairDiffusion's architecture, which is identical to Stable Diffusion's, with the only difference being our adaptive loss rescaling mechanism. During inference, both models exhibit equivalent computational complexity, as verified by the FLOPS and parameter counts presented in Table~\ref{tab:flops_parameters}.}

\begin{table}[H] 
    \centering
    \caption{Comparison of Stable Diffusion and Fair Diffusion on training time per sample}
    \begin{tabular}{lcc}
    \toprule
        & Stable Diffusion  & Fair Diffusion   \\ 
        \midrule
        Training Time / Sample (millisecond)           & 405         & 473           \\ 
        \bottomrule
    \end{tabular}

    \label{tab:training_time}
\end{table}

\begin{table}[H] 
    \centering
    \caption{Computational and Parameter Analysis of Model Components}
    \begin{tabular}{lcccc}
    \toprule
 \multirow{2}{*}{Component}& \multicolumn{2}{c}{Stable Diffusion}& \multicolumn{2}{c}{Fair Diffusion}\\
        
               & FLOPs (GFLOPs) & Parameters (M)  & FLOPs (GFLOPs) &Parameters (M)  \\ 
        \midrule
        UNet            & 675.57         & 865.91          & 675.57         &865.91          \\
        VAE             & 3568.53          & 83.65           & 3568.53          &83.65           \\
        Text Encoder    & 44.61           & 340.39          & 44.61           &340.39          \\ 
        \bottomrule
    \end{tabular}

    \label{tab:flops_parameters}
\end{table}



\end{document}